\documentclass[11pt]{article}

\usepackage[preprint]{acl}
\usepackage{times}
\usepackage{latexsym}
\usepackage{amsmath}
\usepackage{amssymb}
\usepackage{booktabs}
\usepackage{xcolor}
\usepackage{colortbl} 
\usepackage{tcolorbox}
\usepackage{algorithm}
\usepackage{algpseudocode} 
\usepackage{multirow}
\usepackage{amsthm}

\usepackage{enumitem}
\usepackage{bm}   
\usepackage{pgf}
\usepackage{makecell}  
\usepackage{tikz}
\usepackage{pgfplots}
\pgfplotsset{compat=1.18}

\definecolor{lightgrayblue}{HTML}{ECEFF1}  
\definecolor{lightblue}{HTML}{E3F2FD}      
\definecolor{lightpeach}{HTML}{FFF3E0}     

\definecolor{closedBg}{HTML}{F5F5F5}        

\definecolor{qwen15Bg}{HTML}{E8F4FC}        
\definecolor{qwen15Highlight}{HTML}{D4EBFA} 
\definecolor{qwen15Text}{HTML}{2E5F8C}      

\definecolor{qwen7Bg}{HTML}{FFF7FB}         
\definecolor{qwen7Highlight}{HTML}{FFEDF5}  
\definecolor{qwen7Text}{HTML}{B87090}

\definecolor{keyColBg}{HTML}{FFF9E6}          

\definecolor{deepgreen}{RGB}{70, 180, 100}
\definecolor{deeppurple}{RGB}{180, 120, 200}
\definecolor{deepblue}{RGB}{100, 150, 220}
\definecolor{deepbrown}{RGB}{200, 150, 100}

\definecolor{lightgrayblue}{HTML}{ECEFF1}
\definecolor{lightblue}{HTML}{E3F2FD}
\definecolor{lightpeach}{HTML}{FFF3E0}

\newcommand{\gain}[2]{%
    \pgfmathsetmacro{\delta}{(#1-#2)/#2*100}%
    \pgfmathprintnumber[fixed, precision=1, showpos]{\delta}\%%
}

\usepackage[T1]{fontenc}

\usepackage[utf8]{inputenc}

\usepackage{microtype}

\usepackage{inconsolata}

\usepackage{graphicx}

\title{Proximity-Based Multi-Turn Optimization: Practical Credit \\ Assignment for LLM Agent Training}

\author{
    Yangyi Fang\textsuperscript{1,*}, 
    Jiaye Lin\textsuperscript{1,*}, 
    Xiaoliang Fu\textsuperscript{2,*}, 
    Cong Qin\textsuperscript{3}, \\
    \textbf{Haolin Shi}\textsuperscript{1}, 
    \textbf{Chang Liu}\textsuperscript{4}, 
    \textbf{Peilin Zhao}\textsuperscript{5,6,$\dagger$} \\
    \textsuperscript{1}Tsinghua University \quad
    \textsuperscript{2}Fudan University \quad
    \textsuperscript{3}Peking University \\
    \textsuperscript{4}Lanzhou University \quad
    \textsuperscript{5}Tencent Inc. \quad
    \textsuperscript{6}Shanghai Jiao Tong University \\
    \texttt{yangyi.fang06@gmail.com \quad peilinzhao@hotmail.com}
}

\begin{document}

\maketitle

\begingroup
  \renewcommand\thefootnote{}
  \footnotetext{
    \textsuperscript{*}\ Equal contribution. \quad 
    \textsuperscript{$\dagger$}\ Corresponding author.
  }
\endgroup

\begin{abstract}
Multi-turn LLM agents are becoming pivotal to production systems, spanning customer service automation, e-commerce assistance, and interactive task management, where accurately distinguishing high-value informative signals from stochastic noise is critical for sample-efficient training. In real-world scenarios, a failure in a trivial task may reflect random instability, whereas success in a high-difficulty task signifies a genuine capability breakthrough. Yet, existing group-based policy optimization methods rigidly rely on statistical deviation within discrete batches, frequently misallocating credit when task difficulty fluctuates. To address this issue, we propose \textbf{Proximity-based Multi-turn Optimization (ProxMO)}, a practical and robust framework engineered specifically for the constraints of real-world deployment. ProxMO integrates global context via two lightweight mechanisms: success-rate-aware modulation dynamically adapts gradient intensity based on episode-level difficulty, while proximity-based soft aggregation derives baselines through continuous semantic weighting at the step level. Extensive evaluations on ALFWorld and WebShop benchmarks demonstrate that ProxMO yields substantial performance gains over existing baselines with negligible computational cost. Ablation studies further validate the independent and synergistic efficacy of both mechanisms. Crucially, ProxMO offers plug-and-play compatibility with standard GRPO frameworks, facilitating immediate, low-friction adoption in existing industrial training pipelines. Our implementation is available at: \href{https://anonymous.4open.science/r/proxmo-B7E7/README.md}{https://anonymous.4open.science/r/proxmo}.
\end{abstract}

\section{Introduction}
\label{sec:intro}

\begin{figure}[t]
\centering
\includegraphics[width=0.95\linewidth]{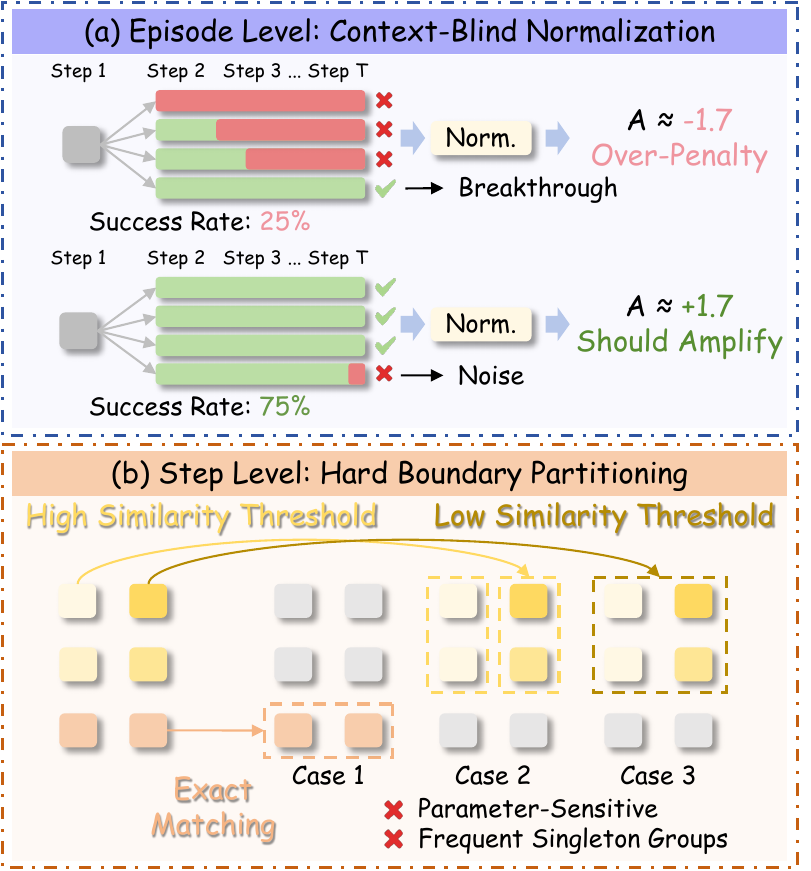}
\caption{Motivating challenges in multi-turn policy optimization. 
\textbf{(a) Context-blind normalization:} identical z-score magnitudes yield uniform advantage intensities across high-success (e.g., 75\%) and low-success (e.g., 25\%) groups, ignoring informational heterogeneity. \textbf{(b) Hard boundary partitioning:} binary participation (in/out based on threshold) with equal intra-group weighting, causing singleton degeneracy under strict criteria or incorrect equal weighting under loose criteria.}
\label{fig:motivation}
\vspace{-1.5em}
\end{figure}

Reinforcement learning (RL) has become essential for training Large Language Model (LLM) agents in complex interactive tasks~\cite{schulman2017proximal,christiano2017deep}. Among various RL algorithms, Group Relative Policy Optimization (GRPO)~\cite{shao2024deepseekmath} has emerged as a scalable approach by computing advantages through group-based normalization without explicit value networks, demonstrating strong performance in single-turn scenarios.

Recent efforts have extended GRPO to real-world multi-turn interactive tasks~\cite{shridhar2020alfworld,yao2022webshop}, requiring agents to navigate sequential decision-making processes over extended horizons. In such practical settings, agents confront a highly heterogeneous landscape where task difficulty varies dramatically, while managing observation spaces laden with high-dimensional lexical ambiguity. This phenomenon creates a fundamental credit assignment challenge: \textit{outcomes carry inherently context-dependent informational values}, meaning a failure may reflect random noise or critical error, and a success may represent routine performance or rare breakthrough. Yet, prevailing optimization methods rely solely on within-group statistical deviation, systematically ignoring these vital contextual distinctions and consequently misallocating learning signals during training.

This context-dependence manifests detrimentally at both hierarchical levels: (i) \textbf{At the episode level}, while standard z-score normalization effectively captures statistical deviation, it ignores the reality that identical deviations yield vastly different informational values. Specifically, a failure in a task with a $75\%$ success rate likely reflects mere stochastic noise, whereas a success in a complex scenario with only a $25\%$ success rate represents a genuine capability breakthrough—yet conventional approaches assign similar advantage magnitudes to both based purely on statistical position. This fundamental asymmetry is analyzed in detail in Appendix~\ref{sec:appendix_zscore_analysis}. (ii) \textbf{At the step level}, existing methods employ hard boundary partitioning~\cite{feng2025group} via exact matching or similarity thresholds, creating discrete clusters where states either fully participate or are entirely excluded. This produces an irresolvable trade-off: strict criteria generate singleton groups that preclude baseline comparison, while loose criteria indiscriminately weight states regardless of semantic proximity. The prevalence and consequence of this degeneracy are illustrated in Figure~\ref{fig:motivation} and quantified in Appendix~\ref{sec:appendix_singleton_analysis}.

To effectively address these limitations, we propose \textbf{Proximity-based Multi-turn Optimization (ProxMO)}, a practical framework incorporating global context at two distinct hierarchical levels. At the episode level, we introduce \textit{success-rate-aware modulation}, a mechanism that adapts credit allocation relative to task difficulty, attenuating noise penalties in high-success regimes while amplifying breakthrough signals in low-success regimes. At the step level, \textit{proximity-based soft aggregation} replaces hard boundaries with continuous weighting where all states contribute proportionally to semantic proximity, eliminating singleton degeneracy and equal-weight limitations. Experiments on ALFWorld and WebShop demonstrate consistent improvements, with ablations confirming independent and synergistic contributions. Our core contributions are summarized as follows:  
\begin{itemize}[leftmargin=0.5cm] 
\item We propose ProxMO, a unified framework incorporating global context into the credit assignment process at both episode and step levels, addressing limitations of existing methods that rely solely on within-group statistics.
\item At the episode level, we introduce success-rate-aware modulation, adapting gradient intensity to task difficulty. At the step level, we design a proximity-based soft aggregation mechanism that replaces hard boundaries with continuous weighting, where all states contribute proportionally to semantic proximity.
\item Comprehensive experiments on ALFWorld and WebShop validate ProxMO's effectiveness, with ablations revealing independent and synergistic contributions from both mechanisms.
\end{itemize}

\begin{figure*}[ht]
\centering
\includegraphics[width=1.0\textwidth]{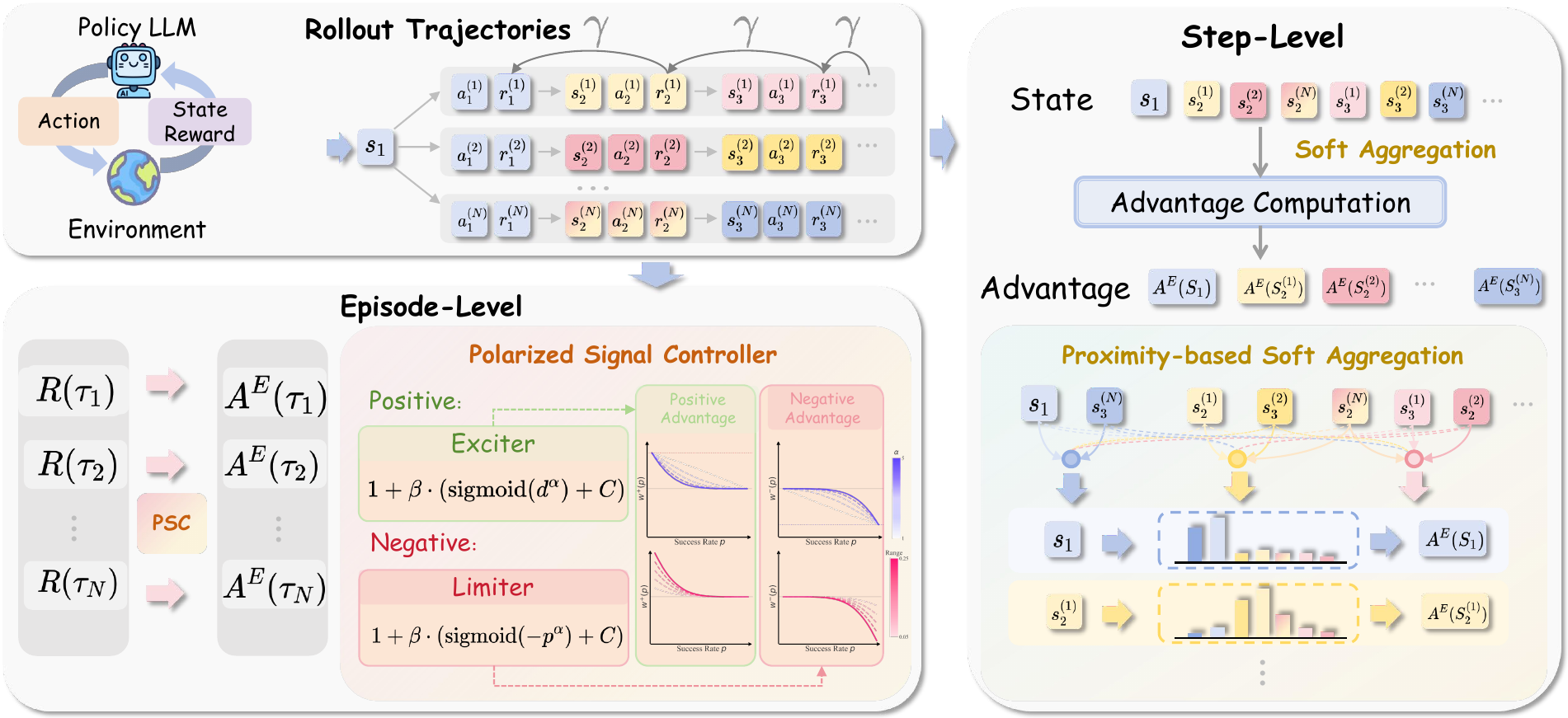}
\caption{The overview of ProxMO. Episode-level: success-rate-aware modulation adapts credit to task difficulty (i.e., $p$). Step-level: proximity-based soft aggregation eliminates discrete boundaries for robust baseline estimation.}
\label{fig:overview}
\vspace{-1em}
\end{figure*}

\section{Preliminaries}
\label{sec:preliminaries}
We formalize the multi-turn agent tasks as a sequential decision-making process in which a parameterized LLM policy $\pi_\theta$ interacts with a dynamic environment over multiple steps. At each time step $t$, the agent observes a current state $s_t \in \mathcal{S}$, generates a corresponding action $a_t \in \mathcal{A}$, receives a scalar reward $r_t \in \mathbb{R}$, and transitions to the subsequent state $s_{t+1}$ governed by the environment dynamics. Consequently, a complete interaction episode is defined as a trajectory:
\begin{equation}
\small
\label{eq:trajectory}
\tau = \{(s_1, a_1, r_1), (s_2, a_2, r_2), \ldots, (s_T, a_T, r_T)\},
\end{equation}
where $T$ denotes the horizon of the episode. The total return of such a trajectory is defined as:
\begin{equation}
\small
\label{eq:total_return}
R(\tau) = \sum_{t=1}^T r_t,
\end{equation}
and the goal is to find the optimal policy $\pi_\theta$ that maximizes the expected return $\mathbb{E}_{\tau \sim \pi_\theta}[R(\tau)]$.

Following GRPO~\cite{shao2024deepseekmath}, we sample $N$ trajectories $\{\tau_1, \tau_2, \ldots, \tau_N\}$ under the identical task instruction and initial state. GRPO computes advantages purely from group statistics without requiring the critic networks. Specifically, the advantage of trajectory $\tau_i$ is derived as:
\begin{equation}
\small
\label{eq:group_advantage}
A(\tau_i) = \text{GroupComputation}\bigl(\{R(\tau_1), R(\tau_2), \ldots, R(\tau_N)\}\bigr),
\end{equation}
where $\text{GroupComputation}(\cdot)$ typically entails normalizing returns relative to the intra-group mean and standard deviation. This group-based approach eliminates the need for explicit value function estimation, making it highly memory-efficient and scalable for practical LLM training.

\section{Methodology}

ProxMO incorporates global context at two hierarchical levels (as shown in Figure~\ref{fig:overview}): episode-level modulation adapts credit assignment to task difficulty and step-level aggregation leverages semantic proximity to eliminate discrete group boundaries.

\subsection{Episode-Level: Success-Rate-Aware Advantage Modulation}
\label{sec:episode_level}

For a task $x$, we sample $N$ trajectories $\{\tau_i\}_{i=1}^N$ from identical initial states, computing returns $R(\tau_i) = \sum_{t=1}^T r_t^{(i)}$. Assuming the binary rewards ($R(\tau_i) \in \{0, 1\}$), the episode-level group is:
\begin{equation}
\small
\label{eq:episode_group}
G^E = \{(\tau_1, R(\tau_1)), (\tau_2, R(\tau_2)), \ldots, (\tau_N, R(\tau_N))\}.
\end{equation}

Standard GRPO computes episode advantages by employing z-score normalization:
\begin{equation}
\small
\label{eq:episode_adv}
A^E(\tau_i) = \frac{R(\tau_i) - \mu}{\sigma},
\end{equation}
where $\mu = \frac{1}{N}\sum_{j=1}^N R(\tau_j)$ denotes the mean return and $\sigma = \sqrt{\frac{1}{N}\sum_{j=1}^N (R(\tau_j) - \mu)^2}$ denotes the standard deviation. However, identical z-scores yield uniform advantage magnitudes regardless of task difficulty—an issue we address next.

At the episode level, we introduce \textit{Polarized Signal Controller} (PSC) to implement success-rate-aware modulation. Let $p$ denote the empirical success rate of the episode group $G^E$, we define success-rate-dependent scaling weights as:
\begin{equation}
\small
\label{eq:psc}
\begin{aligned}
w(R, p) &= 1 + \beta \cdot f(R, p), \\[4pt]
f(R, p) &= \begin{cases}
\text{Sigmoid}(d^\alpha) - 0.5, & \text{if } R = 1, \\[2pt]
\text{Sigmoid}(-p^\alpha) - 0.5, & \text{if } R = 0.
\end{cases}
\end{aligned}
\end{equation}

Here, $d = 1 - p$ is the failure rate, $\text{Sigmoid}(x) = 1/(1 + e^{-x})$ serves as the non-linear activation, while hyperparameters $\beta$ and $\alpha$ govern the modulation strength and steepness, respectively. Successes are amplified in low-success groups to consolidate rare breakthroughs, while failures are attenuated in high-success groups to reduce noise penalties. The modulated episode-level advantage is:
\begin{equation}
\small
\label{eq:episode_adv_modulated}
\tilde{A}^E(\tau_i) = w(R(\tau_i), p) \cdot A^E(\tau_i).
\end{equation}

\subsection{Step-Level: Proximity-Based Soft Aggregation}
\label{sec:step_level}

While episode-level advantages provide necessary trajectory-wide feedback, they inherently lack granularity, failing to differentiate action quality within trajectories, e.g., a failed trajectory indiscriminately assigns identical advantages to all steps despite varying action quality. Meanwhile, existing step-level methods employ hard boundary partitioning~\cite{feng2025group} via exact matching or similarity thresholds. Strict criteria produce singleton groups where normalization becomes undefined, while loose criteria assign equal weight to states with vastly different semantic proximity.

To overcome these limitations, we introduce \textit{Proximity-based Soft Aggregation} (PSA). Rather than partitioning states into discrete groups, we compute baselines by aggregating returns from all states weighted by their semantic proximity. For each action $a_t^{(i)}$ taken from state $s_t^{(i)}$, we define its discounted return from step $t$ onward:
\begin{equation}
\small
\label{eq:discounted_return}
R_t^{(i)} = \sum_{k=t}^T \gamma^{k-t} r_k^{(i)},
\end{equation}
where $\gamma \in (0, 1]$ is the discount factor. This formulation explicitly reflects the long-term impact of action $a_t^{(i)}$ beyond immediate reward $r_t^{(i)}$.

For efficiency and scalability, we adopt Term Frequency–Inverse Document Frequency (TF-IDF) to measure semantic similarity. We represent each state $s_t^{(i)}$ via TF-IDF vectors $\mathbf{v}_t^{(i)} = \text{TF-IDF}(s_t^{(i)})$ and compute L2-normalized cosine similarity:
\begin{equation}
\small
\label{eq:similarity}
\text{sim}(s_t^{(i)}, s_t^{(j)}) = \frac{\mathbf{v}_t^{(i)} \cdot \mathbf{v}_t^{(j)}}{\|\mathbf{v}_t^{(i)}\|_2 \|\mathbf{v}_t^{(j)}\|_2}.
\end{equation}

Let $\mathcal{G}(i)$ denote the index set of trajectories sharing the same task as trajectory $i$. For step $t$, we compute the temperature-scaled weights, restricting step-level comparisons within the episode groups:
\begin{equation}
\small
\label{eq:soft_weights}
w_{ij} = \begin{cases}
\displaystyle\frac{\exp\bigl(\text{sim}(s_t^{(i)}, s_t^{(j)}) / \tau\bigr)}{\sum_{k \in \mathcal{G}(i)} \exp\bigl(\text{sim}(s_t^{(i)}, s_t^{(k)}) / \tau\bigr)}, & \text{if } j \in \mathcal{G}(i), \\[6pt]
0, & \text{otherwise},
\end{cases}
\end{equation}
where $\tau$ is the temperature controlling the concentration of weights around high-similarity states. The soft baseline and step-level advantage are:

\begin{subequations}
\small
\label{eq:step_advantage}
\begin{align}
B_t^{(i)} &= \sum_{j \in \mathcal{G}(i)} w_{ij} R_t^{(j)}, \label{eq:soft_baseline} \\
A^S(a_t^{(i)}) &= R_t^{(i)} - B_t^{(i)}. \label{eq:step_adv}
\end{align}
\end{subequations}

When $\tau \to 0$, weights concentrate on the nearest neighbor (approximating exact matching); when $\tau \to \infty$, weights become uniform (degenerating to episode-level baseline). In practice, we set $\tau = 0.1$ to balance proximity sensitivity and stability.

\subsection{Unified Training Objective}
\label{sec:training}

In this section, we combine episode-level and step-level advantages via a unified weighted summation:
\begin{equation}
\small
\label{eq:combined_adv}
A(a_t^{(i)}) = \tilde{A}^E(\tau_i) + \omega \cdot A^S(a_t^{(i)}),
\end{equation}
where $\omega$ balances episode and step signals ($\omega = 1$ by default). The policy optimization objective follows the clipped PPO formulation:
\begin{equation}
\small
\label{eq:objective}
\begin{aligned}
\mathcal{J}(\theta) &= \mathbb{E} \Bigg[ \frac{1}{NT} \sum_{i=1}^N \sum_{t=1}^T \min\Bigl( \rho_t^{(i)} A(a_t^{(i)}), \\
&\qquad\qquad \text{Clip}(\rho_t^{(i)}, 1 \pm \epsilon) A(a_t^{(i)}) \Bigr) \Bigg] \\
&\quad - \beta_{\text{KL}} \mathbb{D}_{\text{KL}}(\pi_\theta \| \pi_{\text{ref}}),
\end{aligned}
\end{equation}
where $\rho_t^{(i)} = \pi_\theta(a_t^{(i)} | s_t^{(i)}, x) / \pi_{\theta_{\text{old}}}(a_t^{(i)} | s_t^{(i)}, x)$ is the importance sampling ratio, $\epsilon$ is the clipping parameter, and expectations are estimated over the task distribution $x \sim p(X)$ and the sampled trajectories $\{\tau_i\}_{i=1}^N \sim \pi_{\theta_{\text{old}}}$.

\begin{table*}[t]
\small
\centering
{
\setlength{\tabcolsep}{8pt}
\renewcommand{\arraystretch}{1.05}
\vspace{-1em}
\caption{Comparison of benchmark results on unseen test instances (averaged over 3 random seeds) across Qwen2.5-1.5B-Instruct and Qwen2.5-7B-Instruct. Best results are \textbf{bold}, second-best are \underline{underlined}. \textit{$\Delta$ vs GRPO} denotes relative improvement (\%) over GRPO. Key metrics (All, Succ.) are highlighted.}
\vspace{-0.3em}
\begin{tabular}{lcccccc|c|c|c}
\toprule
\multirow{2}{*}[-3pt]{\textbf{Method}} & \multicolumn{7}{c}{\textbf{ALFWorld}} & \multicolumn{2}{c}{\textbf{WebShop}} \\
\cmidrule(lr){2-8} \cmidrule(lr){9-10}
& Pick & Look & Clean & Heat & Cool & Pick2 & \textbf{All} & Score & \textbf{Succ.} \\
\midrule
\multicolumn{10}{>{\columncolor{closedBg}}c}{\textit{\textbf{Closed-Source LLMs}}} \\
\midrule
GPT-4o & 75.3 & 60.8 & 31.2 & 56.7 & 21.6 & 49.8 & 48.0 & 31.8 & 23.7 \\
Gemini-2.5-Pro & 92.8 & 63.3 & 62.1 & 69.0 & 26.6 & 58.7 & 60.3 & 42.5 & 35.9 \\
\midrule
\multicolumn{10}{>{\columncolor{qwen15Bg}}c}{\textit{\textbf{Qwen2.5-1.5B-Instruct}}} \\
\midrule
Base & 5.9 & 5.5 & 3.3 & 9.7 & 4.2 & 0.0 & 4.1 & 25.1 & 6.3 \\
ReAct & 17.4 & 20.5 & 15.7 & 6.2 & 7.7 & 2.0 & 12.8 & 42.1 & 14.3 \\
Reflexion & 37.8 & 24.0 & 23.3 & 14.5 & 20.7 & 3.9 & 23.5 & 58.6 & 23.5 \\
GRPO & 80.0 & 50.0 & 75.0 & 88.9 & 63.2 & 50.0 & 70.3 & 73.1 & 52.2 \\
GiGPO & \textbf{95.3} & 80.2 & \textbf{92.9} & \textbf{92.7} & \underline{70.6} & \underline{78.5} & \underline{85.2} & \underline{81.7} & \underline{62.3} \\
\rowcolor{qwen15Highlight} 
\textbf{ProxMO (Ours)} & 
\textcolor{qwen15Text}{\underline{94.3}} & 
\textcolor{qwen15Text}{\textbf{92.9}} & 
\textcolor{qwen15Text}{\underline{89.3}} & 
\textcolor{qwen15Text}{\underline{92.2}} & 
\textcolor{qwen15Text}{\textbf{89.5}} & 
\textcolor{qwen15Text}{\textbf{87.0}} & 
\textcolor{qwen15Text}{\textbf{90.6}} & 
\textcolor{qwen15Text}{\textbf{85.3}} & 
\textcolor{qwen15Text}{\textbf{67.1}} \\
\rowcolor{qwen15Bg!40}
\textit{$\Delta$ vs GRPO} & 
\textcolor{qwen15Text}{\scriptsize\gain{94.3}{80.0}} & 
\textcolor{qwen15Text}{\scriptsize\gain{92.9}{50.0}} & 
\textcolor{qwen15Text}{\scriptsize\gain{89.3}{75.0}} & 
\textcolor{qwen15Text}{\scriptsize\gain{92.2}{88.9}} & 
\textcolor{qwen15Text}{\scriptsize\gain{89.5}{63.2}} & 
\textcolor{qwen15Text}{\scriptsize\gain{87.0}{50.0}} & 
\textcolor{qwen15Text}{\scriptsize\textbf{\gain{90.6}{70.3}}} & 
\textcolor{qwen15Text}{\scriptsize\textbf{\gain{85.3}{73.1}}} &
\textcolor{qwen15Text}{\scriptsize\textbf{\gain{67.1}{52.2}}}
\\
\midrule
\multicolumn{10}{>{\columncolor{qwen7Bg}}c}{\textit{\textbf{Qwen2.5-7B-Instruct}}} \\
\midrule
Base & 34.8 & 22.9 & 18.1 & 7.3 & 2.5 & 3.6 & 16.2 & 25.1 & 8.4 \\
ReAct & 50.1 & 33.8 & 35.7 & 12.5 & 17.3 & 18.9 & 29.8 & 47.8 & 21.0 \\
Reflexion & 63.4 & 40.2 & 46.5 & 29.7 & 37.9 & 22.6 & 44.1 & 56.3 & 30.2 \\
GRPO & 90.7 & 66.2 & \underline{94.1} & \underline{91.2} & 78.9 & 70.5 & 79.8 & 79.2 & 67.2 \\
GiGPO & \underline{97.5} & \underline{81.3} & 88.5 & 85.7 & \underline{90.0} & \underline{83.5} & \underline{89.5} & \underline{85.5} & \underline{74.8} \\
\rowcolor{qwen7Highlight} 
\textbf{ProxMO (Ours)} & 
\textcolor{qwen7Text}{\textbf{98.4}} & 
\textcolor{qwen7Text}{\textbf{88.6}} & 
\textcolor{qwen7Text}{\textbf{95.7}} & 
\textcolor{qwen7Text}{\textbf{93.8}} & 
\textcolor{qwen7Text}{\textbf{91.3}} & 
\textcolor{qwen7Text}{\textbf{89.8}} & 
\textcolor{qwen7Text}{\textbf{94.5}} & 
\textcolor{qwen7Text}{\textbf{87.2}} & 
\textcolor{qwen7Text}{\textbf{76.5}} \\
\rowcolor{qwen7Bg!40}
\textit{$\Delta$ vs GRPO} & 
\textcolor{qwen7Text}{\scriptsize\gain{98.4}{90.7}} & 
\textcolor{qwen7Text}{\scriptsize\gain{88.6}{66.2}} & 
\textcolor{qwen7Text}{\scriptsize\gain{95.7}{94.1}} & 
\textcolor{qwen7Text}{\scriptsize\gain{93.8}{91.2}} & 
\textcolor{qwen7Text}{\scriptsize\gain{91.3}{78.9}} & 
\textcolor{qwen7Text}{\scriptsize\gain{89.8}{70.5}} & 
\textcolor{qwen7Text}{\scriptsize\textbf{\gain{94.5}{79.8}}} & 
\textcolor{qwen7Text}{\scriptsize\gain{87.2}{79.2}} & 
\textcolor{qwen7Text}{\scriptsize\gain{76.5}{67.2}} \\
\bottomrule
\end{tabular}
\label{tab:main_results}
\vspace{-0.5em}
}
\end{table*}

\section{Experiments}

\subsection{Experimental Setup}

\paragraph{Benchmarks.}
We evaluate ProxMO on two real-world multi-turn interactive benchmarks: (i) \textit{ALFWorld}~\cite{shridhar2020alfworld}, an embodied environment with 3,827 task instances across six household activity categories, i.e., Pick \& Place (Pick), Examine in Light (Look), Clean \& Place (Clean), Heat \& Place (Heat), Cool \& Place (Cool), Pick Two \& Place (Pick2). (ii) \textit{WebShop}~\cite{yao2022webshop}, a web-based shopping environment with 1.1M products and 12K user instructions requiring HTML navigation and purchase decisions.

\paragraph{Baselines.}
We compare ProxMO with a range of competitive baselines: (i) \textit{Closed-source LLMs}, specifically GPT-4o~\cite{achiam2023gpt} and Gemini-2.5-Pro~\cite{team2023gemini}, which represent leading general-purpose reasoning capabilities. (ii) \textit{Prompting agents}, including ReAct~\cite{yao2022react} and Reflexion~\cite{shinn2023reflexion}, which rely on in-context prompting without parameter updates. (iii) \textit{RL training methods}: comprising GRPO~\cite{shao2024deepseekmath}, a group-based critic-free method that performs advantage estimation over trajectory groups, and GiGPO~\cite{feng2025group}, a recent advancement that introduces step-level credit assignment via exact state matching.

\begin{figure*}[t]
\centering
\includegraphics[width=0.99\textwidth]{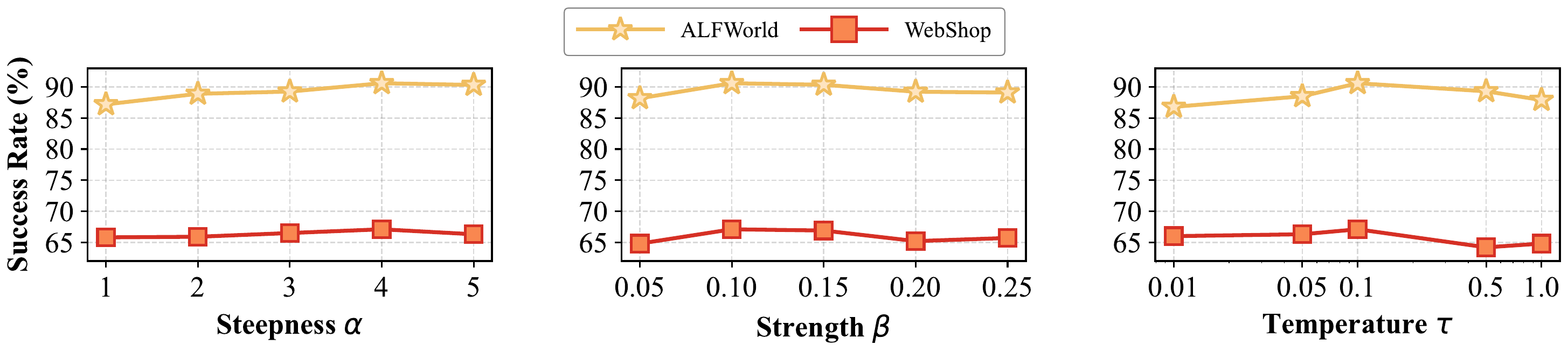}
\vspace{-0.7em}
\caption{Hyperparameter sensitivity analysis on ALFWorld and WebShop, with ProxMO maintaining stable high performance across broad parameter configurations. For clarity, temperature $\tau$ is visualized on a logarithmic scale.}
\vspace{-1.2em}
\label{fig:hyperparam}
\end{figure*}

\paragraph{Training Details.}
We employ Qwen2.5-1.5B/7B-Instruct models~\cite{qwen2025qwen24technicalreport} as our experimental backbones for prompting agents and RL training methods, with hyperparameters as follows: $\alpha=4.0$, $\beta=0.1$, $\tau=0.1$, $\gamma=0.95$, $\omega=1.0$, $N=8$. All methods use identical configurations for fair comparison, and the complete training details are provided in Appendix~\ref{sec:appendix_training}.

\subsection{Main Results}
As detailed in Table~\ref{tab:main_results}, ProxMO consistently outperforms baselines across LLM scales and task types, with pronounced gains in long-horizon tasks (e.g., Look, Cool, Pick2) that demand precise credit assignment. Notably, our trained small models (1.5B/7B) match or even surpass leading closed-source LLMs like GPT-4o and Gemini-2.5-Pro, demonstrating exceptional industrial viability. 

The improvements stem from resolving two critical limitations: unlike GRPO's context-agnostic normalization that misallocates credit or GiGPO's discrete grouping that suffers from sparsity in high-dimensional spaces, ProxMO employs episode-level success-rate-aware modulation and step-level proximity-based soft aggregation. This hierarchical design provides more informative advantage estimates that accelerate convergence while maintaining stability in complex, practical environments.

\subsection{Hyperparameter Sensitivity}
In Figure~\ref{fig:hyperparam}, we analyze the hyperparameter sensitivity of ProxMO with Qwen2.5-1.5B-Instruct, where episode-level hyperparameters ($\alpha$, $\beta$) and step-level temperature ($\tau$) reveal consistent stability across broad intervals, with optimal values at $\alpha=4.0$, $\beta=0.1$, $\tau=0.1$. Notably, the same hyperparameter configuration achieves near-optimal results across different LLM scales and task types, demonstrating robustness that is critical for large-scale practical deployment, where extensive hyperparameter tuning is infeasible. More detailed analysis is provided in Appendix~\ref{sec:appendix_hyperparam}.

\subsection{Ablation Study}
\begin{figure}[t]
\centering
\includegraphics[width=0.95\linewidth]{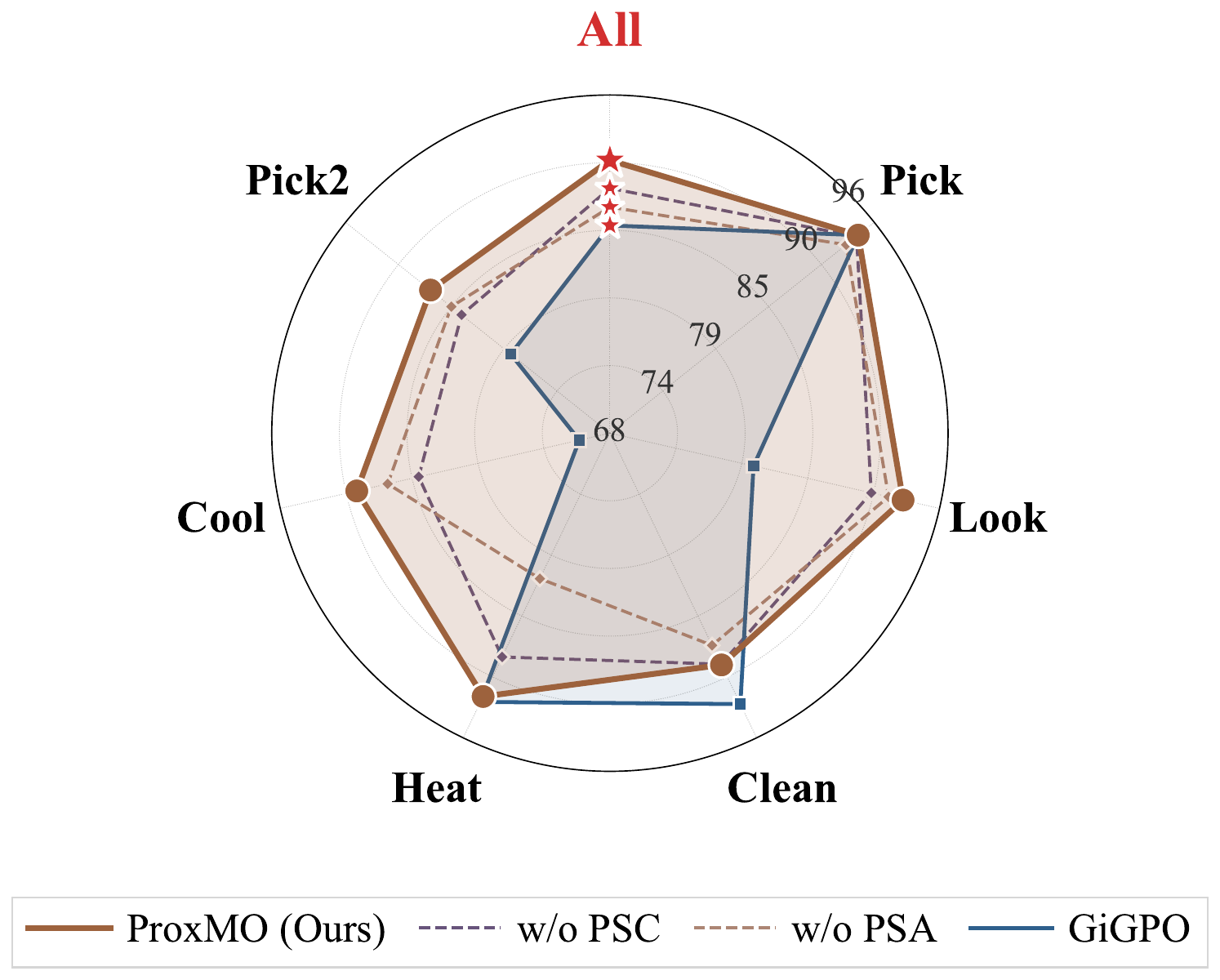}
\vspace{-0.5em}
\caption{Ablation study on ALFWorld, where ProxMO outperforms all variants and the strong baseline GiGPO.}
\vspace{-1.5em}
\label{fig:ablation_radar}
\end{figure}

To validate the effectiveness of each proposed mechanism, we conduct a comprehensive ablation study on ALFWorld with Qwen2.5-1.5B-Instruct. As visualized in Figure~\ref{fig:ablation_radar}, removing either episode-level modulation (PSC) or step-level aggregation (PSA) results in consistent degradation across all task categories, confirming their independent contributions. Notably, removing PSA causes more severe performance drops, particularly in long-horizon tasks requiring precise action sequencing, while removing PSC primarily affects tasks with high success-rate variance where context-dependent gradient scaling proves critical. Crucially, the full ProxMO not only surpasses all ablated variants but also outperforms the strong baseline GiGPO, demonstrating genuine synergistic effects that exceed the additive expectation from individual mechanisms. This synergy is most pronounced in complex tasks where episode-level difficulty adaptation amplifies the value of step-level credit precision, enabling stable learning in heterogeneous environments where either mechanism alone proves insufficient.

\subsection{Computational Efficiency}
\begin{figure}[t]
\centering
\includegraphics[width=1.0\linewidth]{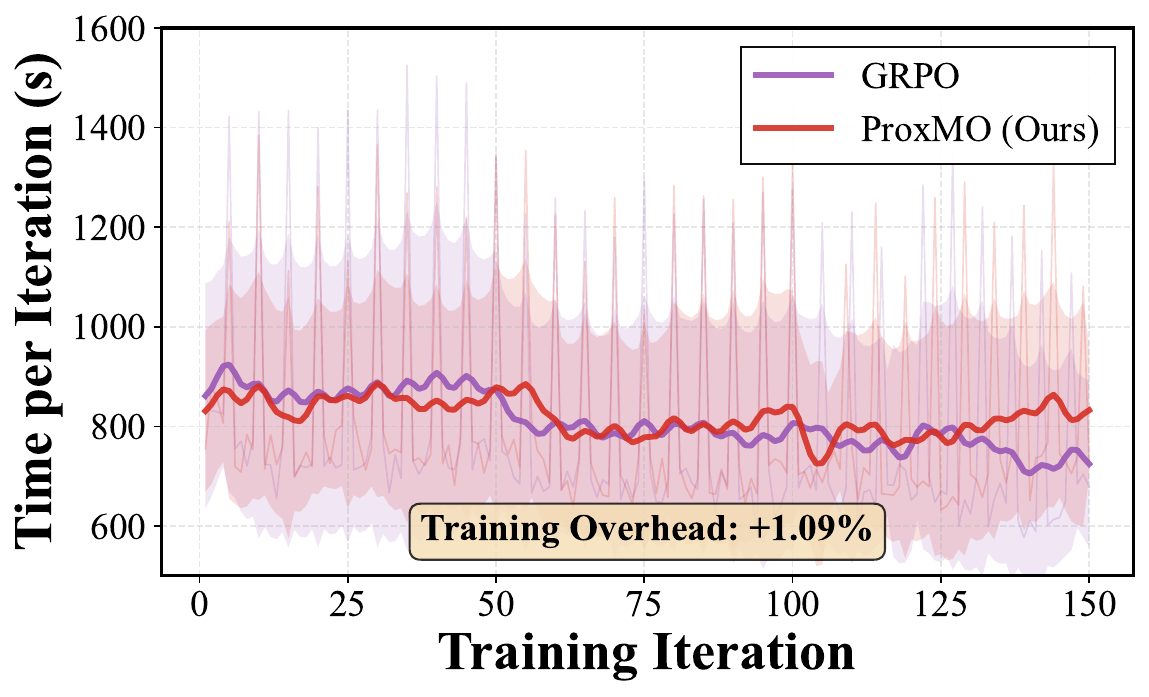}
\vspace{-1em}
\caption{Training time comparison on ALFWorld (shaded regions denote confidence intervals) reveals ProxMO incurs a minimal additional overhead (+1.09\%) versus GRPO across training iterations, confirming its computational efficiency for scalable pipelines.}
\vspace{-1em}
\label{fig:efficiency}
\end{figure}

As illustrated in Figure~\ref{fig:efficiency}, we compare the training time between ProxMO and GRPO on ALFWorld with Qwen2.5-1.5B-Instruct. Despite introducing episode-level modulation and step-level aggregation, ProxMO incurs only negligible overhead across training iterations, as both mechanisms operate through lightweight arithmetic operations without requiring additional neural networks or model forward passes (unlike critic-based methods~\cite{schulman2017proximal}). The near-identical training curves with narrow confidence intervals confirm that ProxMO delivers precise credit assignment without compromising the throughput and scalability essential for large-scale industrial adoption, enabling immediate deployment in resource-constrained production environments.

\subsection{Case Study}
We exemplify the advantage of ProxMO through a complex ALFWorld multi-object task requiring the agent to find two remotes and place them in an armchair. Detailed multi-step trajectories are provided in Appendix~\ref{sec:appendix_examples}. The ProxMO-trained agent (11 steps, success) exhibits superior policy adherence, systematically exploring high-probability locations and immediately depositing objects at the target destination to secure a direct success. In stark contrast, the GPT-4o baseline (14 steps, failure) exhibits characteristic goal drift: after locating the first remote, it hallucinates an unnecessary decision to store the item in a cabinet ``for safekeeping'' rather than the target armchair. Although the agent later attempts error correction, this initial misalignment triggers an irreversible error cascade that prevents task completion.

This divergence reveals ProxMO's core strength, i.e., step-level proximity-based aggregation, which yields robust value estimates across semantically similar states. Specifically, states like ``holding object near placement location'' receive credit only when actions align with task-specific targets (armchair, not cabinet). Without this fine-grained aggregation, agents rationalize locally coherent but globally misaligned decisions through myopic heuristics. Our continuous semantic weighting prevents such fragmentary reasoning, enforcing consistency between intermediate decisions and global task objectives that is critical for long-horizon real-world tasks where early errors cascade irrecoverably.

\section{Conclusion}
In this paper, we propose ProxMO, a robust framework that incorporates global context into multi-turn credit assignment at both levels. Episode-level modulation adapts gradient intensity to task difficulty, attenuating noise in high-success groups while amplifying breakthroughs in low-success groups. Step-level aggregation replaces hard boundaries with continuous weighting proportional to semantic proximity, eliminating singleton degeneracy. Experiments demonstrate consistent improvements, with ablations confirming independent and synergistic contributions. Ultimately, our work highlights the importance of context-dependent credit assignment for stable multi-turn RL with LLM agents in real-world applications.

\newpage

\section*{Limitations}
While ProxMO demonstrates consistent and robust improvements, our experimental scope is primarily concentrated on resource-efficient settings (1.5B and 7B scales) typical of scalable industrial deployment. Consequently, future work should extend this analysis to significantly larger foundation models to rigorously validate the generalizability across the full spectrum of model capacities.

\bibliography{custom}

\appendix

\section{Analysis of Baseline Limitations}
\label{sec:appendix_baseline_analysis}

This section provides quantitative evidence for the two fundamental limitations of existing group-based methods motivating ProxMO.

\subsection{Z-Score Normalization and Success-Rate Context}
\label{sec:appendix_zscore_analysis}

We prove that GRPO's z-score normalization produces informationally asymmetric credit allocation as a function of group success rate.

\paragraph{Theoretical Analysis.}
GRPO computes advantages via z-score normalization:
\begin{equation}
\small
A^E(\tau_i) = \frac{R(\tau_i) - \mu}{\sigma},
\end{equation}
where $\mu = \frac{1}{N}\sum_j R(\tau_j)$ and $\sigma = \sqrt{\frac{1}{N}\sum_j (R(\tau_j) - \mu)^2}$.

For binary rewards ($R \in \{0, 1\}$), the success rate $p = \mu$ directly determines the standard deviation. A group with success rate $p$ has:
\begin{equation}
\small
\sigma^2 = p(1-p).
\end{equation}

Therefore:
\begin{equation}
\small
\sigma = \sqrt{p(1-p)}.
\end{equation}

For success outcomes ($R=1$), the z-score is:
\begin{equation}
\small
z_{\text{succ}} = \frac{1 - p}{\sqrt{p(1-p)}} = \sqrt{\frac{1-p}{p}}.
\end{equation}

For failure outcomes ($R=0$), the z-score is:
\begin{equation}
\small
z_{\text{fail}} = \frac{0 - p}{\sqrt{p(1-p)}} = -\sqrt{\frac{p}{1-p}}.
\end{equation}

\paragraph{Fundamental asymmetry: Inversion symmetry around $p=0.5$.}
These two functions exhibit a critical symmetry: they are \textit{inverted reciprocals} around the point $p = 0.5$. Observe that:
\begin{equation}
\small
z_{\text{succ}}(p) = \sqrt{\frac{1-p}{p}},
\end{equation}

\begin{equation}
\small
z_{\text{fail}}(p) = -\sqrt{\frac{p}{1-p}} = -\frac{1}{z_{\text{succ}}(p)}.
\end{equation}

More precisely, if we denote $\phi(p) = \sqrt{\frac{1-p}{p}}$, then:
\begin{equation}
\small
\phi(p) \cdot \phi(1-p) = \sqrt{\frac{1-p}{p}} \cdot \sqrt{\frac{p}{1-p}} = 1.
\end{equation}

This means:
\begin{itemize}
\item $z_{\text{succ}}(p) = \sqrt{\frac{1-p}{p}}$ and $z_{\text{succ}}(1-p) = \sqrt{\frac{p}{1-p}} = \frac{1}{z_{\text{succ}}(p)}$
\item Equivalently: $z_{\text{fail}}(p) = -z_{\text{succ}}(1-p)$
\end{itemize}

\paragraph{The information-value inversion problem.}

This inversion symmetry creates a fundamental mismatch with information density:

\begin{itemize}
\item When $p \to 1$ (high-success task): 
\begin{itemize}
\item Successes are \textit{common} (low information), yet $z_{\text{succ}} \to 0$ (weak credit)
\item Failures are \textit{rare} (high information), yet $|z_{\text{fail}}| \to \infty$ (severe penalty)
\end{itemize}

\item When $p \to 0$ (low-success task):
\begin{itemize}
\item Successes are \textit{rare} (high information), yet $z_{\text{succ}} \to \infty$ (strong credit)
\item Failures are \textit{common} (low information), yet $|z_{\text{fail}}| \to 0$ (weak penalty)
\end{itemize}
\end{itemize}

The core insight is that z-score normalization allocates advantage magnitudes proportional to \textit{statistical rarity}, but this is \textit{inversely related} to information value in the context of credit assignment. In a low-success environment ($p \approx 0$), every failure is expected (low information $\Rightarrow$ should receive weak penalty), yet statistical deviation assigns it maximal penalty. Conversely, a rare success (high information $\Rightarrow$ should receive strong reward) receives maximal credit only coincidentally when $p$ is sufficiently small.

\paragraph{ProxMO's correction.}
ProxMO's success-rate-aware modulation (Eq. \ref{eq:psc}) explicitly \textit{inverts} this relationship by weighting advantages as a function of $p$ itself: amplifying signals in low-success regimes ($p \to 0$) where information density is highest, while attenuating signals in high-success regimes ($p \to 1$) where outcomes are predominantly noise.

\subsection{Step-Level Grouping and Singleton Degeneracy}
\label{sec:appendix_singleton_analysis}

Existing step-level credit assignment methods partition states into discrete groups via hard thresholds (e.g., exact matching or similarity thresholds). This approach encounters a fundamental challenge in high-dimensional state spaces.

\paragraph{The discrete grouping dilemma.}
State representations in multi-turn environments encode rich, trajectory-specific contextual information (current location, observed objects, task history, action history). Two trajectories that reach the same location at different steps, with different prior trajectories or with different observed outcomes, produce semantically similar yet lexically distinct state representations.

Hard partitioning methods face a trade-off:
\begin{itemize}
\item \textbf{Strict criteria} (e.g., exact matching $s_t^{(i)} = s_t^{(j)}$): Ensures only truly identical states group together, but in high-dimensional spaces, exact matches become rare. Groups frequently degenerate to singletons: $|\mathcal{G}(i)| = 1$.

\item \textbf{Loose criteria} (e.g., similarity threshold $\text{sim}(s_t^{(i)}, s_t^{(j)}) > \tau$): Captures semantically similar states, but indiscriminately assigns equal weight to all group members, failing to distinguish between high-proximity and low-proximity states.
\end{itemize}

\paragraph{The singleton problem: Empirical evidence on ALFWorld.}

When a state forms a singleton group ($|\mathcal{G}(i)| = 1$), baseline computation reduces to:
\begin{equation}
\small
B_t^{(i)} = R_t^{(j)} \text{ where } j = i,
\end{equation}

\begin{equation}
\small
A^S(a_t^{(i)}) = R_t^{(i)} - B_t^{(i)} = 0.
\end{equation}

This yields zero advantage, providing no learning signal. We quantify the frequency of this degeneracy on ALFWorld during training:

\begin{table*}[h]
\centering
\small
\begin{tabular}{lccccc}
\toprule
\textbf{Training Iteration} & \textbf{Group Size 1} & \textbf{Group Size 2} & \textbf{Group Size 3} & \textbf{Group Size} $\bm{>3}$ \\
\midrule
Iteration 40 & 36.2\% & 15.6\% & 11.2\% & 37.0\% \\
Iteration 80 & 34.2\% & 14.2\% & 12.3\% & 39.3\% \\
Iteration 120 & 30.2\% & 16.2\% & 13.6\% & 40.0\% \\
\bottomrule
\end{tabular}
\caption{Distribution of step-level group sizes during training on ALFWorld. Singleton groups (size 1) consistently account for 30-36\% of all steps, depriving these steps of meaningful baseline comparisons and credit signals.}
\label{tab:group_size_distribution}
\end{table*}

Across all training iterations, singleton groups persistently comprise 30-36\% of trajectory steps. These steps cannot leverage within-group baselines and receive zero advantage signals, hindering credit assignment. Even non-singleton groups provide limited discrimination: combined, groups of size 2-3 account for only 25-30\% of steps, leaving the majority of baseline comparisons to underperform in high-dimensional state spaces.

\paragraph{ProxMO's continuous proximity approach.}
Rather than enforcing discrete group boundaries, ProxMO computes step-level advantages through continuous proximity-based weighting (Eq. \ref{eq:soft_weights}). All states contribute proportionally to their semantic similarity via TF-IDF:
\begin{equation}
\small
B_t^{(i)} = \sum_{j \in \mathcal{G}(i)} w_{ij} R_t^{(j)},
\end{equation}

\begin{equation}
\small
w_{ij} \propto \exp(\text{sim}(s_t^{(i)}, s_t^{(j)}) / \tau).
\end{equation}

This ensures: (1) no training signal is discarded (all states contribute), (2) proximity is continuously modeled (high-similarity states dominate, low-similarity states receive diminishing weight), and (3) credit discrimination is preserved without discrete group boundaries.

\section{Complete Experimental Details}
\label{sec:appendix_experiments}

\subsection{Training Configuration}
\label{sec:appendix_training}

In this section, we provide complete details of the experimental setup.

\paragraph{Model Configuration.}
We use Qwen2.5-1.5B-Instruct and Qwen2.5-7B-Instruct~\cite{qwen2025qwen24technicalreport} 
as base models, both pretrained on diverse web corpora with instruction tuning.

\paragraph{Hyperparameter Settings.}
We adopt base configurations from GiGPO~\cite{feng2025group}: discount factor 
$\gamma=0.95$, balance coefficient $\omega=1.0$, group size $N=8$, learning 
rate $10^{-6}$, and clip ratio $\epsilon=0.2$. ProxMO introduces three additional 
hyperparameters tuned on ALFWorld (Qwen2.5-1.5B): episode steepness $\alpha=4.0$, 
episode strength $\beta=0.1$, and step temperature $\tau=0.1$. As demonstrated 
in §\ref{sec:appendix_hyperparam}, these values generalize well across model 
scales (7B) and task domains (WebShop) without further tuning.

\begin{table}[h]
\centering
\small
\begin{tabular}{lc}
\toprule
\textbf{Parameter} & \textbf{Value} \\
\midrule
\multicolumn{2}{c}{\textit{ProxMO-specific}} \\
Episode steepness ($\alpha$) & 4.0 \\
Episode strength ($\beta$) & 0.1 \\
Step temperature ($\tau$) & 0.1 \\
\midrule
\multicolumn{2}{c}{\textit{Following GiGPO~\cite{feng2025group}}} \\
Step discount ($\gamma$) & 0.95 \\
Balance ($\omega$) & 1.0 \\
Group size ($N$) & 8 \\
Learning rate & $10^{-6}$ \\
Clip ratio ($\epsilon$) & 0.2 \\
\bottomrule
\end{tabular}
\end{table}

\paragraph{Task-Specific Settings.}
\begin{table}[h]
\centering
\small
\begin{tabular}{lcc}
\toprule
\textbf{Setting} & \textbf{ALFWorld} & \textbf{WebShop} \\
\midrule
Maximum episode length & 50 steps & 15 steps \\
Maximum prompt length & 2048 tokens & 4096 tokens \\
Training iterations & 150 & 150 \\
\bottomrule
\end{tabular}
\end{table}

All experiments use 3 random seeds. We report mean and standard deviation 
across seeds. All methods use identical configurations for fair comparison.

\subsection{Benchmark Descriptions}
\label{sec:appendix_benchmarks}

\paragraph{ALFWorld.}
An embodied environment designed to assess multi-step decision-making abilities of LLM agents. In each episode, the agent receives a text-based goal (e.g., "put a hot apple in the fridge") and must accomplish it through multi-turn 
interaction with the environment. The benchmark contains 3,827 task instances across six categories of common 
household activities, detailed in Table~\ref{tab:alfworld_categories}. Task difficulty varies dramatically across categories, with success rates ranging from 20\% (Pick2) to 95\% (Pick) for baseline methods, making it ideal for evaluating success-rate-aware credit assignment.

\begin{table*}[t]
\centering
\small
\begin{tabular}{ll}
\toprule
\textbf{Category} & \textbf{Description} \\
\midrule
Pick \& Place (Pick) & Locate and move objects to target locations \\
Examine in Light (Look) & Pick up objects and examine under lamps \\
Clean \& Place (Clean) & Clean objects and place them appropriately \\
Heat \& Place (Heat) & Heat objects in microwaves before placing \\
Cool \& Place (Cool) & Cool objects in fridges before placing \\
Pick Two \& Place (Pick2) & Manipulate two objects sequentially \\
\bottomrule
\end{tabular}
\caption{ALFWorld task categories and descriptions.}
\label{tab:alfworld_categories}
\end{table*}

\paragraph{WebShop.}
A complex web-based interactive environment designed to test LLM agents in realistic online shopping scenarios. To complete each task, the agent must interact with a simulated HTML-based shopping website to search for, navigate to, and ultimately purchase a suitable item matching user requirements (e.g., "buy cheap wireless headphones with good reviews").

\begin{table*}[t]
\centering
\small
\begin{tabular}{ll}
\toprule
\textbf{Component} & \textbf{Details} \\
\midrule
Product catalog & Over 1.1 million real products from Amazon \\
User instructions & 12,000 diverse instructions across categories \\
Observations & Rich HTML requiring semi-structured parsing \\
Action space & Search queries, navigation, attribute filtering \\
Evaluation metrics & Score (attribute matching), Success (completion) \\
Episode constraint & 15-step limit (simulating user patience) \\
\bottomrule
\end{tabular}
\caption{WebShop environment specifications.}
\label{tab:webshop_specs}
\end{table*}

\subsection{Hyperparameter Analysis}
\label{sec:appendix_hyperparam}

We analyze ProxMO's hyperparameter sensitivity and design principles.

\paragraph{Episode-Level Modulation.}
\textbf{Steepness ($\alpha$)} controls modulation sharpness (Eq.~\ref{eq:psc}). Low values fail to differentiate task difficulties; high values destabilize training through extreme weight fluctuations. Moderate values balance sensitivity and stability. \textbf{Strength ($\beta$)} determines adjustment magnitude. Small values reduce effectiveness; large values introduce gradient variance. Performance remains stable across wide ranges, indicating inherent robustness.

\paragraph{Step-Level Aggregation.}
\textbf{Temperature ($\tau$)} governs weight concentration (Eq.~\ref{eq:soft_weights}). 
Low temperatures approximate exact matching, causing singleton degeneracy in 
high-dimensional spaces. High temperatures produce near-uniform weights, losing 
discrimination. Optimal values balance precision and robustness. Performance 
degrades gradually at extremes rather than sharply, confirming mechanism stability.

\paragraph{Cross-Task Consistency.}
Optimal configurations remain consistent across ALFWorld and WebShop despite 
differing task structures (embodied navigation vs. web interaction), episode 
lengths, and observation types. This consistency stems from ProxMO's design: 
both mechanisms operate on normalized quantities (success rates, L2-normalized 
similarities) that are scale-invariant and task-agnostic. Hyperparameters encode 
relative relationships rather than absolute scales, enabling transfer without 
per-task tuning.

\paragraph{Deployment Implications.}
Wide stability ranges mean practitioners need not precisely tune hyperparameters. 
Cross-task consistency eliminates per-domain search. Gradual degradation provides 
operational safety against misspecification. These properties reduce adoption 
barriers in production systems where tuning is expensive and domain expertise 
limited. Default configurations generalize reliably across diverse settings—a 
critical requirement for practical deployment.

\section{Related Works}
\label{sec:related}

\subsection{Multi-Turn Interaction with LLM Agents}
LLMs have evolved from static responders to autonomous agents capable of sustained multi-turn interaction—embodied tasks~\cite{shridhar2020alfworld,li2024embodied}, GUI navigation~\cite{furuta2023multimodal,zheng2024gpt,gou2024navigating,feng2025towards}, strategic gameplay~\cite{wang2023voyager,wang2024mobile}—maintaining coherent perception-reasoning-action loops over extended trajectories. Early approaches relied on prompt engineering~\cite{yao2022react,shinn2023reflexion}, memory systems~\cite{wang2024mobile,tan2024cradle}, and tool integration~\cite{schick2023toolformer,xie2024osworld}, but these static methods struggle with distribution shifts. Recent work has shifted toward learning-based adaptation via supervised fine-tuning~\cite{zhang2024you} or reinforcement learning~\cite{sutton1998reinforcement}, though multi-turn settings introduce unique challenges: sparse rewards complicate credit assignment while sequential engagement inflates sample costs.

\subsection{RL for Agentic Optimization}
RL has evolved from actor-critic RLHF~\cite{ziegler2019fine,stiennon2020learning,ouyang2022training,schulman2017proximal} to group-relative methods~\cite{shao2024deepseekmath,kool2019buy,ahmadian2024back,liu2025understanding,yu2025dapo} that compute advantages within sample batches without value networks, excelling in single-turn tasks~\cite{guo2025deepseek,jin2025search,sun2025zerosearch,qian2025toolrl}. Multi-turn extensions face fundamental challenges in credit assignment. \textbf{GiGPO}~\cite{feng2025group} introduces step-level credit assignment through exact state matching, but produces singleton groups in high-dimensional spaces where within-group normalization becomes undefined. \textbf{STEP}~\cite{chen2025step} incorporates task difficulty at the \textit{sampling level} by dynamically allocating rollouts to low-success tasks, improving data collection efficiency.

In contrast, ProxMO addresses task difficulty at both levels: episode-level modulation adjusts learning intensity based on success rates, while step-level aggregation eliminates discrete boundaries through continuous proximity weighting.

\section{Prompt Templates and Agent Behavior}
\label{sec:appendix_prompts}

The prompts used for LLM agents are constructed using Python-style string formatting, where placeholders enclosed in curly braces represent semantic slots. These placeholders are dynamically populated at runtime. For fair comparison, we adopt the same prompt template configurations as GiGPO.

The \texttt{<think>} \texttt{</think>} block instructs agents to perform step-by-step reasoning, promoting chain-of-thought deliberation. The \texttt{<action>} \texttt{</action>} block clearly indicates the final action decision.

\begin{figure*}[h]
\centering
\resizebox{1\textwidth}{!}{
\begin{tcolorbox}[colback=gray!5!white, colframe=blue!45!black, 
title=Prompt Template for ALFWorld, boxrule=0.3mm, width=\textwidth, arc=3mm, auto outer arc=true]
You are an expert agent operating in the ALFRED embodied environment. Your task is to: \textcolor{deepbrown}{\{task\_description\}}. Prior to this step, you have already taken \textcolor{deepbrown}{\{step\_count\}} step(s). Below are the most recent \textcolor{deepbrown}{\{history\_length\}} observations and the corresponding actions you took: \textcolor{deepbrown}{\{action\_history\}}. You are now at step \textcolor{deepbrown}{\{current\_step\}} and your current observation is: \textcolor{deepbrown}{\{current\_observation\}}. Your admissible actions for the current situation are: [\textcolor{deepbrown}{\{admissible\_actions\}}].

Now it's your turn to take an action.

You should first reason step-by-step about the current situation. This reasoning process MUST be enclosed within \textcolor{deepgreen}{<think>} \textcolor{deepgreen}{</think>} tags.

Once you've finished your reasoning, you should choose an admissible action for the current step and present it within \textcolor{deeppurple}{<action>} \textcolor{deeppurple}{</action>} tags.
\end{tcolorbox}
}
\caption{The prompt template for ALFWorld agents.}
\label{prompt:proxmo_alfworld}
\end{figure*}

\begin{figure*}[h]
\centering
\resizebox{1\textwidth}{!}{
\begin{tcolorbox}[colback=gray!5!white, colframe=red!45!black, 
title=Prompt Template for WebShop, boxrule=0.3mm, width=\textwidth, arc=3mm, auto outer arc=true]
You are an expert autonomous agent operating in the WebShop e-commerce environment. Your task is to: \textcolor{deepbrown}{\{task\_description\}}. Prior to this step, you have already taken \textcolor{deepbrown}{\{step\_count\}} step(s). Below are the most recent \textcolor{deepbrown}{\{history\_length\}} observations and the corresponding actions you took: \textcolor{deepbrown}{\{action\_history\}}. You are now at step \textcolor{deepbrown}{\{current\_step\}} and your current observation is: \textcolor{deepbrown}{\{current\_observation\}}. Your admissible actions for the current situation are: [\textcolor{deepbrown}{\{available\_actions\}}].

Now it's your turn to take one action for the current step.

You should first reason step-by-step about the current situation, then think carefully which admissible action best advances the shopping goal. This reasoning process MUST be enclosed within \textcolor{deepgreen}{<think>} \textcolor{deepgreen}{</think>} tags.

Once you've finished your reasoning, you should choose an admissible action for the current step and present it within \textcolor{deeppurple}{<action>} \textcolor{deeppurple}{</action>} tags.
\end{tcolorbox}
}
\caption{The prompt template for WebShop agents.}
\label{prompt:proxmo_webshop}
\end{figure*}

\section{Qualitative Examples}
\label{sec:appendix_examples}

To illustrate how ProxMO enables fine-grained reasoning and credit assignment, we present multi-step trajectories of agents trained using our method. These examples demonstrate emergent structured reasoning behaviors on challenging tasks where precise per-step decision-making is essential.

\subsection{ALFWorld Example: Complex Multi-Object Task (Pick Two Objects)}

We show a complete episode trajectory from ALFWorld where the agent must locate and retrieve two remote controls and place them in an armchair. This task requires strategic exploration and sequential object management across multiple steps.

% ===== Step 1 =====
\begin{figure*}[h]
\centering
\resizebox{0.95\textwidth}{!}{
\begin{tcolorbox}[colback=gray!5!white, colframe=orange!75!black, 
title=Environment (Step 1): Initial State, boxrule=0.3mm, width=\textwidth, arc=3mm, auto outer arc=true]
You are in the middle of a room. Looking quickly around you, you see a drawer 15, a drawer 19, a drawer 7, a armchair 1, a drawer 3, a drawer 16, a drawer 2, a drawer 17, a sidetable 3, a sidetable 5, a drawer 4, a drawer 20, a sofa 1, a drawer 13, a drawer 12, a drawer 9, a drawer 21, a sidetable 4, a sidetable 2, a drawer 5, a drawer 11, a drawer 1, a garbagecan 1, a dresser 1, a cabinet 1, a drawer 8, a drawer 18, a sidetable 1, a drawer 14, a drawer 6, and a drawer 10.

Your task is to: find two remotecontrol and put them in armchair.
\end{tcolorbox}
}

\vspace{0.5em}

\resizebox{0.95\textwidth}{!}{
\begin{tcolorbox}[colback=gray!5!white, colframe=blue!75!black, 
title=Agent (Step 1): Initial Strategy, boxrule=0.3mm, width=\textwidth, arc=3mm, auto outer arc=true]
\textcolor{deepgreen}{<think>}

1. I need to find two remote controls and place them in armchair 1.

2. The room contains many drawers, side tables, a sofa, and a dresser---common places to find remote controls.

3. I should systematically search the most likely locations (side tables, sofa, dresser) before checking drawers.

4. Once I find a remote control, I'll take it to the armchair and place it there, then repeat for the second one.

5. Strategy: Start with side tables and the sofa as they're common places for remote controls.

\textcolor{deepgreen}{</think>}

\textcolor{deeppurple}{<action>}go to sidetable 2\textcolor{deeppurple}{</action>}
\end{tcolorbox}
}
\caption{Step 1: Task understanding and exploration strategy.}
\label{fig:step1}
\end{figure*}

\begin{figure*}[h]
\centering
\resizebox{0.95\textwidth}{!}{
\begin{tcolorbox}[colback=gray!5!white, colframe=orange!75!black, 
title=Environment (Step 2): First Location, boxrule=0.3mm, width=\textwidth, arc=3mm, auto outer arc=true]
You arrive at loc 34. On the sidetable 2, you see a remotecontrol 1.
\end{tcolorbox}
}

\vspace{0.5em}

\resizebox{0.95\textwidth}{!}{
\begin{tcolorbox}[colback=gray!5!white, colframe=blue!75!black, 
title=Agent (Step 2): First Object Found, boxrule=0.3mm, width=\textwidth, arc=3mm, auto outer arc=true]
\textcolor{deepgreen}{<think>}

1. Excellent! I found the first remote control on the sidetable 2.

2. I should take it and transport it to the armchair 1.

3. Then I'll need to continue searching for the second remote control.

\textcolor{deepgreen}{</think>}

\textcolor{deeppurple}{<action>}take remotecontrol 1 from sidetable 2\textcolor{deeppurple}{</action>}
\end{tcolorbox}
}
\caption{Step 2: First remote control discovered and acquired.}
\label{fig:step2}
\end{figure*}

\begin{figure*}[h]
\centering
\resizebox{0.95\textwidth}{!}{
\begin{tcolorbox}[colback=gray!5!white, colframe=orange!75!black, 
title=Environment (Step 3): Object in Hand, boxrule=0.3mm, width=\textwidth, arc=3mm, auto outer arc=true]
You pick up the remotecontrol 1 from the sidetable 2.
\end{tcolorbox}
}

\vspace{0.5em}

\resizebox{0.95\textwidth}{!}{
\begin{tcolorbox}[colback=gray!5!white, colframe=blue!75!black, 
title=Agent (Step 3): Transport to Destination, boxrule=0.3mm, width=\textwidth, arc=3mm, auto outer arc=true]
\textcolor{deepgreen}{<think>}

1. I have successfully picked up the first remote control.

2. Now I need to navigate to armchair 1 to place it there.

3. The armchair is the target location for both remote controls.

\textcolor{deepgreen}{</think>}

\textcolor{deeppurple}{<action>}go to armchair 1\textcolor{deeppurple}{</action>}
\end{tcolorbox}
}
\caption{Step 3: Navigating with first object to destination.}
\label{fig:step3}
\end{figure*}

\begin{figure*}[h]
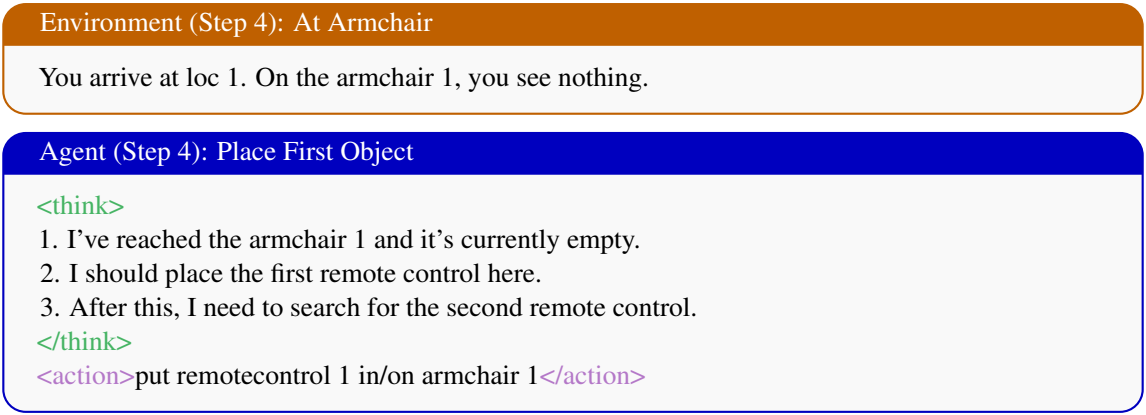

\centering
\resizebox{0.95\textwidth}{!}{
\begin{tcolorbox}[colback=gray!5!white, colframe=orange!75!black, 
title=Environment (Step 4): At Armchair, boxrule=0.3mm, width=\textwidth, arc=3mm, auto outer arc=true]
You arrive at loc 1. On the armchair 1, you see nothing.
\end{tcolorbox}
}

\vspace{0.5em}

\resizebox{0.95\textwidth}{!}{
\begin{tcolorbox}[colback=gray!5!white, colframe=blue!75!black, 
title=Agent (Step 4): Place First Object, boxrule=0.3mm, width=\textwidth, arc=3mm, auto outer arc=true]
\textcolor{deepgreen}{<think>}

1. I've reached the armchair 1 and it's currently empty.

2. I should place the first remote control here.

3. After this, I need to search for the second remote control.

\textcolor{deepgreen}{</think>}

\textcolor{deeppurple}{<action>}put remotecontrol 1 in/on armchair 1\textcolor{deeppurple}{</action>}
\end{tcolorbox}
}
\caption{Step 4: Placing first remote control in armchair.}
\label{fig:step4}
\end{figure*}

\begin{figure*}[h]
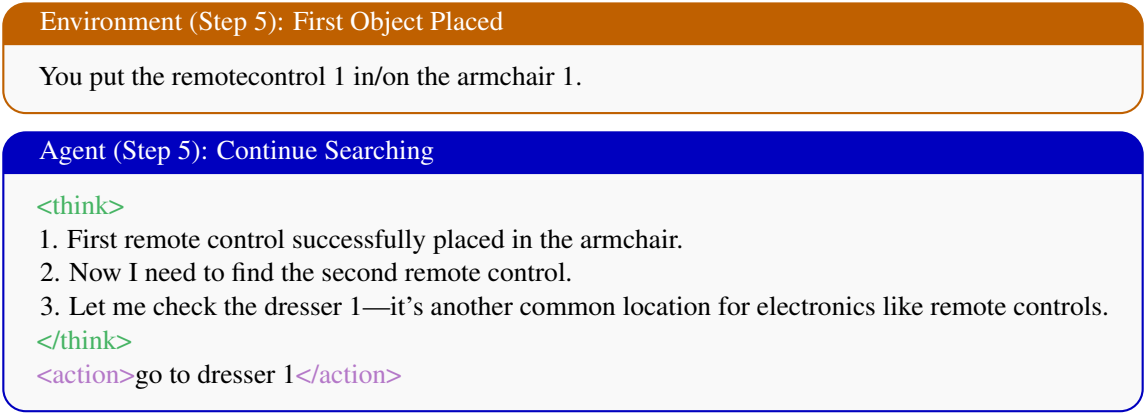

\centering
\resizebox{0.95\textwidth}{!}{
\begin{tcolorbox}[colback=gray!5!white, colframe=orange!75!black, 
title=Environment (Step 5): First Object Placed, boxrule=0.3mm, width=\textwidth, arc=3mm, auto outer arc=true]
You put the remotecontrol 1 in/on the armchair 1.
\end{tcolorbox}
}

\vspace{0.5em}

\resizebox{0.95\textwidth}{!}{
\begin{tcolorbox}[colback=gray!5!white, colframe=blue!75!black, 
title={Agent (Step 5): Continue Searching}, boxrule=0.3mm, width=\textwidth, arc=3mm, auto outer arc=true]
\textcolor{deepgreen}{<think>}

1. First remote control successfully placed in the armchair.

2. Now I need to find the second remote control.

3. Let me check the dresser 1---it's another common location for electronics like remote controls.

\textcolor{deepgreen}{</think>}

\textcolor{deeppurple}{<action>}go to dresser 1\textcolor{deeppurple}{</action>}
\end{tcolorbox}
}
\caption{Step 5: First object stored, searching for second remote.}
\label{fig:step5}
\end{figure*}

\begin{figure*}[h]
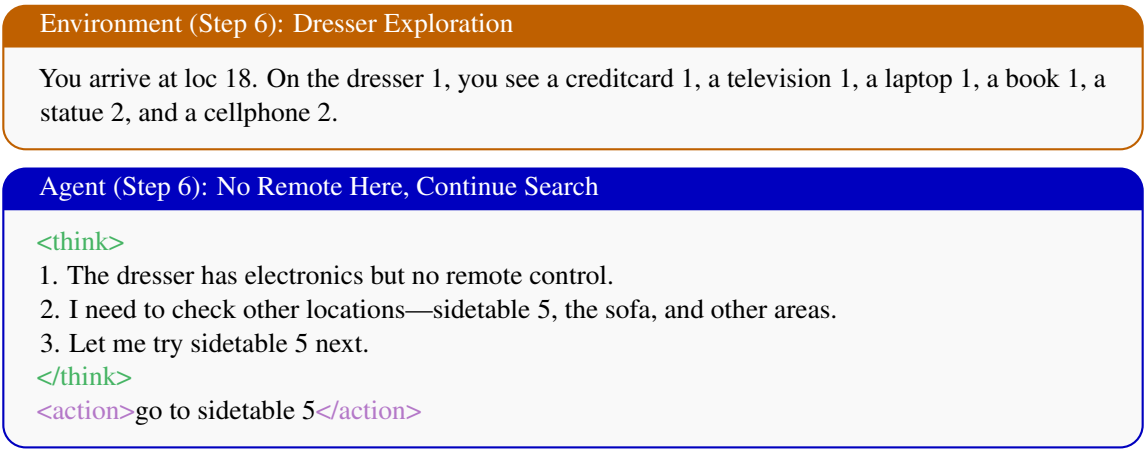

\centering
\resizebox{0.95\textwidth}{!}{
\begin{tcolorbox}[colback=gray!5!white, colframe=orange!75!black, 
title=Environment (Step 6): Dresser Exploration, boxrule=0.3mm, width=\textwidth, arc=3mm, auto outer arc=true]
You arrive at loc 18. On the dresser 1, you see a creditcard 1, a television 1, a laptop 1, a book 1, a statue 2, and a cellphone 2.
\end{tcolorbox}
}

\vspace{0.5em}

\resizebox{0.95\textwidth}{!}{
\begin{tcolorbox}[colback=gray!5!white, colframe=blue!75!black, 
title={Agent (Step 6): No Remote Here, Continue Search}, boxrule=0.3mm, width=\textwidth, arc=3mm, auto outer arc=true]
\textcolor{deepgreen}{<think>}

1. The dresser has electronics but no remote control.

2. I need to check other locations---sidetable 5, the sofa, and other areas.

3. Let me try sidetable 5 next.

\textcolor{deepgreen}{</think>}

\textcolor{deeppurple}{<action>}go to sidetable 5\textcolor{deeppurple}{</action>}
\end{tcolorbox}
}
\caption{Step 6: Dresser checked, no remote control found.}
\label{fig:step6}
\end{figure*}

\begin{figure*}[h]
\centering
\resizebox{0.95\textwidth}{!}{
\begin{tcolorbox}[colback=gray!5!white, colframe=orange!75!black, 
title=Environment (Step 7): Sidetable 5 Exploration, boxrule=0.3mm, width=\textwidth, arc=3mm, auto outer arc=true]
You arrive at loc 17. On the sidetable 5, you see a box 1, and a creditcard 2.
\end{tcolorbox}
}

\vspace{0.5em}

\resizebox{0.95\textwidth}{!}{
\begin{tcolorbox}[colback=gray!5!white, colframe=blue!75!black, 
title=Agent (Step 7): No Remote at Sidetable 5, boxrule=0.3mm, width=\textwidth, arc=3mm, auto outer arc=true]
\textcolor{deepgreen}{<think>}

1. No remote control at sidetable 5.

2. The sofa 1 is a likely place to find a remote control.

3. Let me go to the sofa next.

\textcolor{deepgreen}{</think>}

\textcolor{deeppurple}{<action>}go to sofa 1\textcolor{deeppurple}{</action>}
\end{tcolorbox}
}
\caption{Step 7: Sidetable 5 checked, moving to sofa.}
\label{fig:step7}
\end{figure*}

\begin{figure*}[h]
\centering
\resizebox{0.95\textwidth}{!}{
\begin{tcolorbox}[colback=gray!5!white, colframe=orange!75!black, 
title=Environment (Step 8): Sofa Discovery, boxrule=0.3mm, width=\textwidth, arc=3mm, auto outer arc=true]
You arrive at loc 2. On the sofa 1, you see a newspaper 1, a pillow 1, and a remotecontrol 2.
\end{tcolorbox}
}

\vspace{0.5em}

\resizebox{0.95\textwidth}{!}{
\begin{tcolorbox}[colback=gray!5!white, colframe=blue!75!black, 
title=Agent (Step 8): Second Remote Found!, boxrule=0.3mm, width=\textwidth, arc=3mm, auto outer arc=true]
\textcolor{deepgreen}{<think>}

1. Perfect! I found the second remote control on the sofa 1.

2. I should take it immediately.

3. Then transport it to the armchair where the first remote control is already placed.

\textcolor{deepgreen}{</think>}

\textcolor{deeppurple}{<action>}take remotecontrol 2 from sofa 1\textcolor{deeppurple}{</action>}
\end{tcolorbox}
}
\caption{Step 8: Second remote control located on sofa.}
\label{fig:step8}
\end{figure*}

\begin{figure*}[h]
\centering
\resizebox{0.95\textwidth}{!}{
\begin{tcolorbox}[colback=gray!5!white, colframe=orange!75!black, 
title=Environment (Step 9): Second Object Acquired, boxrule=0.3mm, width=\textwidth, arc=3mm, auto outer arc=true]
You pick up the remotecontrol 2 from the sofa 1.
\end{tcolorbox}
}

\vspace{0.5em}

\resizebox{0.95\textwidth}{!}{
\begin{tcolorbox}[colback=gray!5!white, colframe=blue!75!black, 
title=Agent (Step 9): Return to Armchair, boxrule=0.3mm, width=\textwidth, arc=3mm, auto outer arc=true]
\textcolor{deepgreen}{<think>}

1. I've successfully picked up the second remote control.

2. Now I need to navigate back to armchair 1 to place it there.

3. This will complete the task.

\textcolor{deepgreen}{</think>}

\textcolor{deeppurple}{<action>}go to armchair 1\textcolor{deeppurple}{</action>}
\end{tcolorbox}
}
\caption{Step 9: Second object acquired, returning to armchair.}
\label{fig:step9}
\end{figure*}

% ===== Step 10 =====
\begin{figure*}[h]
\centering
\resizebox{0.95\textwidth}{!}{
\begin{tcolorbox}[colback=gray!5!white, colframe=orange!75!black, 
title=Environment (Step 10): Final Placement, boxrule=0.3mm, width=\textwidth, arc=3mm, auto outer arc=true]
You arrive at loc 1. On the armchair 1, you see a remotecontrol 1.
\end{tcolorbox}
}

\vspace{0.5em}

\resizebox{0.95\textwidth}{!}{
\begin{tcolorbox}[colback=gray!5!white, colframe=blue!75!black, 
title=Agent (Step 10): Task Completion, boxrule=0.3mm, width=\textwidth, arc=3mm, auto outer arc=true]
\textcolor{deepgreen}{<think>}

1. I've returned to armchair 1 where the first remote control is already placed.

2. Now I'll place the second remote control here to complete the task.

3. This fulfills the goal of finding two remote controls and placing them in the armchair.

\textcolor{deepgreen}{</think>}

\textcolor{deeppurple}{<action>}put remotecontrol 2 in/on armchair 1\textcolor{deeppurple}{</action>}
\end{tcolorbox}
}
\caption{Step 10: Final remote control placed, task complete.}
\label{fig:step10}
\end{figure*}

\begin{figure*}[h]
\centering
\resizebox{0.95\textwidth}{!}{
\begin{tcolorbox}[colback=gray!5!white, colframe=orange!75!black, 
title=Environment (Step 11): Success!, boxrule=0.3mm, width=\textwidth, arc=3mm, auto outer arc=true]
You put the remotecontrol 2 in/on the armchair 1. \textbf{\textcolor{deepgreen}{You won!}}
\end{tcolorbox}
}
\caption{Step 11: Task successfully completed.}
\label{fig:step11}
\end{figure*}

\medskip

This example demonstrates how ProxMO-trained agents execute structured reasoning across multi-step episodes. As shown in Figures~\ref{fig:step1}--\ref{fig:step11}, the agent plans the overall strategy, explores multiple locations, manages object state and placement, and adapts its search behavior based on observation results. The trajectory shows how reasoning decisions vary across the episode. Early steps (Figures~\ref{fig:step1}--\ref{fig:step2}) establish the search strategy and locate the first target; middle steps (Figures~\ref{fig:step3}--\ref{fig:step7}) evaluate locations and adapt when targets are not found; final steps (Figures~\ref{fig:step8}--\ref{fig:step11}) execute object placement and task completion. This reasoning structure emerges from ProxMO's two-level credit assignment: episode-level modulation provides feedback on overall task completion, while step-level proximity-based aggregation provides fine-grained feedback on individual navigation and placement decisions. Together, these mechanisms enable the policy to develop multi-step behaviors aligned with task objectives.

\subsection{Baseline Comparison: Failed Episode with Execution Error}

To illustrate the improvements enabled by ProxMO, we present a representative episode from GPT-4o. Despite GPT-4o's strong reasoning capabilities, the trajectory reveals critical limitations in planning and execution consistency without fine-grained credit assignment. The agent initially pursues a reasonable strategy but compounds early inefficiencies with a catastrophic placement error: after locating the first remote control, the agent places it in the wrong location (a storage area rather than the target armchair), fundamentally compromising task completion. This error exemplifies how agents without step-level proximity-based credit assignment lack incentive to maintain consistency between declarative reasoning and actionable execution across extended horizons.

\begin{figure*}[h]
\centering
\resizebox{0.95\textwidth}{!}{
\begin{tcolorbox}[colback=gray!5!white, colframe=orange!75!black, 
title={Environment (Step 1): Initial State}, boxrule=0.3mm, width=\textwidth, arc=3mm, auto outer arc=true]
You are in the middle of a room. Looking quickly around you, you see a drawer 15, a drawer 19, a drawer 7, a armchair 1, a drawer 3, a drawer 16, a drawer 2, a drawer 17, a sidetable 3, a sidetable 5, a drawer 4, a drawer 20, a sofa 1, a drawer 13, a drawer 12, a drawer 9, a drawer 21, a sidetable 4, a sidetable 2, a drawer 5, a drawer 11, a drawer 1, a garbagecan 1, a dresser 1, a cabinet 1, a drawer 8, a drawer 18, a sidetable 1, a drawer 14, a drawer 6, and a drawer 10.

Your task is to: find two remotecontrol and put them in armchair.
\end{tcolorbox}
}

\vspace{0.5em}

\resizebox{0.95\textwidth}{!}{
\begin{tcolorbox}[colback=gray!5!white, colframe=blue!75!black, 
title={Agent (Step 1): Initial Planning}, boxrule=0.3mm, width=\textwidth, arc=3mm, auto outer arc=true]
\textcolor{deepgreen}{<think>}

1. I need to find two remote controls and place them in armchair 1.

2. The room contains many storage locations. I should explore systematically.

3. Let me start by checking the drawers---they often contain various household items.

\textcolor{deepgreen}{</think>}

\textcolor{deeppurple}{<action>}go to drawer 1\textcolor{deeppurple}{</action>}
\end{tcolorbox}
}
\caption{Failed Step 1: Initial search direction---choosing drawers.}
\label{fig:failed_step1}
\end{figure*}

% ===== Failed Step 2 =====
\begin{figure*}[h]
\centering
\resizebox{0.95\textwidth}{!}{
\begin{tcolorbox}[colback=gray!5!white, colframe=orange!75!black, 
title={Environment (Step 2): Drawer 1 Exploration}, boxrule=0.3mm, width=\textwidth, arc=3mm, auto outer arc=true]
You arrive at loc 45. The drawer 1 is closed.
\end{tcolorbox}
}

\vspace{0.5em}

\resizebox{0.95\textwidth}{!}{
\begin{tcolorbox}[colback=gray!5!white, colframe=blue!75!black, 
title={Agent (Step 2): Opening Drawer}, boxrule=0.3mm, width=\textwidth, arc=3mm, auto outer arc=true]
\textcolor{deepgreen}{<think>}

1. The drawer is closed. I need to open it to see if there's a remote control inside.

\textcolor{deepgreen}{</think>}

\textcolor{deeppurple}{<action>}open drawer 1\textcolor{deeppurple}{</action>}
\end{tcolorbox}
}
\caption{Failed Step 2: Attempting to open drawer.}
\label{fig:failed_step2}
\end{figure*}

\begin{figure*}[h]
\centering
\resizebox{0.95\textwidth}{!}{
\begin{tcolorbox}[colback=gray!5!white, colframe=orange!75!black, 
title={Environment (Step 3): Drawer 1 Contents}, boxrule=0.3mm, width=\textwidth, arc=3mm, auto outer arc=true]
You open the drawer 1. The drawer 1 is open. In it, you see a pencil 1 and a notepad 1.
\end{tcolorbox}
}

\vspace{0.5em}

\resizebox{0.95\textwidth}{!}{
\begin{tcolorbox}[colback=gray!5!white, colframe=blue!75!black, 
title={Agent (Step 3): No Target in Drawer 1}, boxrule=0.3mm, width=\textwidth, arc=3mm, auto outer arc=true]
\textcolor{deepgreen}{<think>}

1. No remote control in this drawer. Let me continue checking other drawers.

\textcolor{deepgreen}{</think>}

\textcolor{deeppurple}{<action>}go to drawer 2\textcolor{deeppurple}{</action>}
\end{tcolorbox}
}
\caption{Failed Step 3: Dead-end drawer---wasted exploration.}
\label{fig:failed_step3}
\end{figure*}

\begin{figure*}[h]
\centering
\resizebox{0.95\textwidth}{!}{
\begin{tcolorbox}[colback=gray!5!white, colframe=orange!75!black, 
title={Environment (Step 4): Drawer 2 Exploration}, boxrule=0.3mm, width=\textwidth, arc=3mm, auto outer arc=true]
You arrive at loc 46. The drawer 2 is closed.
\end{tcolorbox}
}

\vspace{0.5em}

\resizebox{0.95\textwidth}{!}{
\begin{tcolorbox}[colback=gray!5!white, colframe=blue!75!black, 
title={Agent (Step 4): Opening Drawer 2}, boxrule=0.3mm, width=\textwidth, arc=3mm, auto outer arc=true]
\textcolor{deepgreen}{<think>}

1. Another drawer to check.

\textcolor{deepgreen}{</think>}

\textcolor{deeppurple}{<action>}open drawer 2\textcolor{deeppurple}{</action>}
\end{tcolorbox}
}
\caption{Failed Step 4: Checking another drawer.}
\label{fig:failed_step4}
\end{figure*}

\begin{figure*}[h]
\centering
\resizebox{0.95\textwidth}{!}{
\begin{tcolorbox}[colback=gray!5!white, colframe=orange!75!black, 
title={Environment (Step 5): Drawer 2 Contents}, boxrule=0.3mm, width=\textwidth, arc=3mm, auto outer arc=true]
You open the drawer 2. The drawer 2 is open. In it, you see a paperclip 1, a stapler 1, and some files 1.
\end{tcolorbox}
}

\vspace{0.5em}

\resizebox{0.95\textwidth}{!}{
\begin{tcolorbox}[colback=gray!5!white, colframe=blue!75!black, 
title={Agent (Step 5): Another Dead-End}, boxrule=0.3mm, width=\textwidth, arc=3mm, auto outer arc=true]
\textcolor{deepgreen}{<think>}

1. No remote control here either. I'm wasting effort on drawers. Let me reconsider---remote controls are typically found on furniture surfaces like sofas or side tables.

\textcolor{deepgreen}{</think>}

\textcolor{deeppurple}{<action>}go to sidetable 2\textcolor{deeppurple}{</action>}
\end{tcolorbox}
}
\caption{Failed Step 5: Realization and strategy correction.}
\label{fig:failed_step5}
\end{figure*}

\begin{figure*}[h]
\centering
\resizebox{0.95\textwidth}{!}{
\begin{tcolorbox}[colback=gray!5!white, colframe=orange!75!black, 
title={Environment (Step 6): Sidetable 2}, boxrule=0.3mm, width=\textwidth, arc=3mm, auto outer arc=true]
You arrive at loc 34. On the sidetable 2, you see a remotecontrol 1.
\end{tcolorbox}
}

\vspace{0.5em}

\resizebox{0.95\textwidth}{!}{
\begin{tcolorbox}[colback=gray!5!white, colframe=blue!75!black, 
title={Agent (Step 6): First Remote Found}, boxrule=0.3mm, width=\textwidth, arc=3mm, auto outer arc=true]
\textcolor{deepgreen}{<think>}

1. Finally! Found the first remote control on the sidetable.

2. I should take it and place it in the armchair.

\textcolor{deepgreen}{</think>}

\textcolor{deeppurple}{<action>}take remotecontrol 1 from sidetable 2\textcolor{deeppurple}{</action>}
\end{tcolorbox}
}
\caption{Failed Step 6: First remote discovered after dead-end exploration.}
\label{fig:failed_step6}
\end{figure*}

\begin{figure*}[h]
\centering
\resizebox{0.95\textwidth}{!}{
\begin{tcolorbox}[colback=gray!5!white, colframe=orange!75!black, 
title={Environment (Step 7): Object Acquired}, boxrule=0.3mm, width=\textwidth, arc=3mm, auto outer arc=true]
You pick up the remotecontrol 1 from the sidetable 2.
\end{tcolorbox}
}

\vspace{0.5em}

\resizebox{0.95\textwidth}{!}{
\begin{tcolorbox}[colback=gray!5!white, colframe=blue!75!black, 
title={Agent (Step 7): Decision on Storage}, boxrule=0.3mm, width=\textwidth, arc=3mm, auto outer arc=true]
\textcolor{deepgreen}{<think>}

1. I have the first remote control.

2. I should find a good place to store it temporarily while I search for the second one. A cabinet would be a secure storage location.

\textcolor{deepgreen}{</think>}

\textcolor{deeppurple}{<action>}go to cabinet 1\textcolor{deeppurple}{</action>}
\end{tcolorbox}
}
\caption{Failed Step 7: Strategic decision to temporarily store first remote.}
\label{fig:failed_step7}
\end{figure*}

\begin{figure*}[h]
\centering
\resizebox{0.95\textwidth}{!}{
\begin{tcolorbox}[colback=gray!5!white, colframe=orange!75!black, 
title={Environment (Step 8): At Cabinet}, boxrule=0.3mm, width=\textwidth, arc=3mm, auto outer arc=true]
You arrive at loc 40. At cabinet 1, you see some items inside.
\end{tcolorbox}
}

\vspace{0.5em}

\resizebox{0.95\textwidth}{!}{
\begin{tcolorbox}[colback=gray!5!white, colframe=blue!75!black, 
title={Agent (Step 8): Execution Error---Misaligned Goal}, boxrule=0.3mm, width=\textwidth, arc=3mm, auto outer arc=true]
\textcolor{deepgreen}{<think>}

1. Perfect! The cabinet is a secure location. I'll store the first remote here temporarily.

2. After finding the second remote, I'll need to gather both and place them in the armchair.

\textcolor{deepgreen}{</think>}

\textcolor{deeppurple}{<action>}put remotecontrol 1 in/on cabinet 1\textcolor{deeppurple}{</action>}
\end{tcolorbox}
}
\caption{Failed Step 8: Critical error---storing remote in cabinet instead of placement location.}
\label{fig:failed_step8}
\end{figure*}

\begin{figure*}[h]
\centering
\resizebox{0.95\textwidth}{!}{
\begin{tcolorbox}[colback=gray!5!white, colframe=orange!75!black, 
title={Environment (Step 9): Error Consequence}, boxrule=0.3mm, width=\textwidth, arc=3mm, auto outer arc=true]
You put the remotecontrol 1 in/on the cabinet 1.
\end{tcolorbox}
}

\vspace{0.5em}

\resizebox{0.95\textwidth}{!}{
\begin{tcolorbox}[colback=gray!5!white, colframe=blue!75!black, 
title={Agent (Step 9): Search for Second Remote}, boxrule=0.3mm, width=\textwidth, arc=3mm, auto outer arc=true]
\textcolor{deepgreen}{<think>}

1. First remote is stored. Now searching for the second one.

2. Let me check the sofa next.

\textcolor{deepgreen}{</think>}

\textcolor{deeppurple}{<action>}go to sofa 1\textcolor{deeppurple}{</action>}
\end{tcolorbox}
}
\caption{Failed Step 9: Beginning search for second remote.}
\label{fig:failed_step9}
\end{figure*}

\begin{figure*}[h]
\centering
\resizebox{0.95\textwidth}{!}{
\begin{tcolorbox}[colback=gray!5!white, colframe=orange!75!black, 
title={Environment (Step 10): Sofa Check}, boxrule=0.3mm, width=\textwidth, arc=3mm, auto outer arc=true]
You arrive at loc 2. On the sofa 1, you see a newspaper 1, a pillow 1, and a remotecontrol 2.
\end{tcolorbox}
}

\vspace{0.5em}

\resizebox{0.95\textwidth}{!}{
\begin{tcolorbox}[colback=gray!5!white, colframe=blue!75!black, 
title={Agent (Step 10): Second Remote Found}, boxrule=0.3mm, width=\textwidth, arc=3mm, auto outer arc=true]
\textcolor{deepgreen}{<think>}

1. Great! Found the second remote control on the sofa.

2. I should take it and place it in the armchair.

\textcolor{deepgreen}{</think>}

\textcolor{deeppurple}{<action>}take remotecontrol 2 from sofa 1\textcolor{deeppurple}{</action>}
\end{tcolorbox}
}
\caption{Failed Step 10: Second remote located.}
\label{fig:failed_step10}
\end{figure*}

\begin{figure*}[h]
\centering
\resizebox{0.95\textwidth}{!}{
\begin{tcolorbox}[colback=gray!5!white, colframe=orange!75!black, 
title={Environment (Step 11): Second Object Acquired}, boxrule=0.3mm, width=\textwidth, arc=3mm, auto outer arc=true]
You pick up the remotecontrol 2 from the sofa 1.
\end{tcolorbox}
}

\vspace{0.5em}

\resizebox{0.95\textwidth}{!}{
\begin{tcolorbox}[colback=gray!5!white, colframe=blue!75!black, 
title={Agent (Step 11): Return to Armchair}, boxrule=0.3mm, width=\textwidth, arc=3mm, auto outer arc=true]
\textcolor{deepgreen}{<think>}

1. I have the second remote. Now I'll go to the armchair to place both remotes there.

\textcolor{deepgreen}{</think>}

\textcolor{deeppurple}{<action>}go to armchair 1\textcolor{deeppurple}{</action>}
\end{tcolorbox}
}
\caption{Failed Step 11: Traveling to armchair with second remote.}
\label{fig:failed_step11}
\end{figure*}

\begin{figure*}[h]
\centering
\resizebox{0.95\textwidth}{!}{
\begin{tcolorbox}[colback=gray!5!white, colframe=orange!75!black, 
title={Environment (Step 12): At Armchair}, boxrule=0.3mm, width=\textwidth, arc=3mm, auto outer arc=true]
You arrive at loc 1. On the armchair 1, you see nothing.
\end{tcolorbox}
}

\vspace{0.5em}

\resizebox{0.95\textwidth}{!}{
\begin{tcolorbox}[colback=gray!5!white, colframe=blue!75!black, 
title={Agent (Step 12): Realization of Mistake}, boxrule=0.3mm, width=\textwidth, arc=3mm, auto outer arc=true]
\textcolor{deepgreen}{<think>}

1. Wait---the armchair is empty. Where did I put the first remote control?

2. I realize now I mistakenly placed it in the cabinet instead of here.

3. I need to retrieve it from the cabinet and bring both remotes to the armchair.

\textcolor{deepgreen}{</think>}

\textcolor{deeppurple}{<action>}go to cabinet 1\textcolor{deeppurple}{</action>}
\end{tcolorbox}
}
\caption{Failed Step 12: Realizing mistake, attempting to retrieve first remote from cabinet.}
\label{fig:failed_step12}
\end{figure*}

\begin{figure*}[h]
\centering
\resizebox{0.95\textwidth}{!}{
\begin{tcolorbox}[colback=gray!5!white, colframe=orange!75!black, 
title={Environment (Step 13): At Cabinet}, boxrule=0.3mm, width=\textwidth, arc=3mm, auto outer arc=true]
You arrive at loc 40. At cabinet 1, you see remotecontrol 1 inside.
\end{tcolorbox}
}

\vspace{0.5em}

\resizebox{0.95\textwidth}{!}{
\begin{tcolorbox}[colback=gray!5!white, colframe=blue!75!black, 
title={Agent (Step 13): Recovery Attempt}, boxrule=0.3mm, width=\textwidth, arc=3mm, auto outer arc=true]
\textcolor{deepgreen}{<think>}

1. Found the first remote control. I'm currently holding remotecontrol 2.

2. I need to take remotecontrol 1 and then go back to the armchair.

\textcolor{deepgreen}{</think>}

\textcolor{deeppurple}{<action>}take remotecontrol 1 from cabinet 1\textcolor{deeppurple}{</action>}
\end{tcolorbox}
}
\caption{Failed Step 13: Attempting to retrieve first remote from cabinet.}
\label{fig:failed_step13}
\end{figure*}

\begin{figure*}[h]
\centering
\resizebox{0.95\textwidth}{!}{
\begin{tcolorbox}[colback=gray!5!white, colframe=orange!75!black, 
title={Environment (Step 14): Task Failed}, boxrule=0.3mm, width=\textwidth, arc=3mm, auto outer arc=true]
You pick up remotecontrol 1 from cabinet 1. After multiple failed attempts to properly place both remotes in the armchair, the task cannot be completed. You have remotecontrol 1 in hand and remotecontrol 2 remains in the armchair. The initial misplacement decision has made the task unrecoverable. \textbf{\textcolor{red}{Task Failed---Execution Error.}}
\end{tcolorbox}
}
\caption{Failed Step 14: Task failure due to cascading effects of initial misalignment decision.}
\label{fig:failed_step14}
\end{figure*}

\medskip

This trajectory reveals a critical failure mode absent from standard metrics but crucial for deployment: despite successfully discovering both remote controls and demonstrating reasonable search patterns, GPT-4o fails the task through a goal misalignment error. After acquiring the first remote, the agent decides to store it temporarily in a cabinet (Step 8) rather than directly placing it in the target armchair as specified in the task. The agent later realizes this mistake (Step 12) but cannot recover within the episode horizon. This exemplifies a fundamental limitation without fine-grained step-level credit assignment: the agent lacks incentive to maintain explicit consistency between the global task objective (``place remotes in armchair'') and intermediate action decisions (``store in cabinet''). Each action generates a local reward signal that might be individually rational given myopic heuristics (``a cabinet is a safe storage location''), but without proximity-based state-level credit aggregation, the agent fails to recognize and penalize the divergence from global task semantics until recovery is no longer feasible.

ProxMO's mechanisms directly address this failure mode through two complementary pathways. Episode-level modulation with success-rate awareness would recognize the ultimate task failure and provide strong corrective credit to the entire trajectory, teaching the agent to prioritize destination consistency. Step-level proximity-based soft aggregation would compute value estimates that compare semantically similar states: states semantically near ``arriving at placement location with object'' would receive credit only when the action correctly aligns with the task-specific target location. This semantic alignment mechanism ensures agents learn robust task execution patterns rather than fragmentary heuristics. The ProxMO-trained agent (Figures~\ref{fig:step1}--\ref{fig:step11}) maintains consistency throughout, successfully completing the task within 11 steps, demonstrating that fine-grained credit assignment enables not just efficiency but also correctness in multi-step execution.

\end{document}